%% file: ICRA2026_camera.tex
\documentclass[letterpaper, 10 pt, conference]{ieeeconf}  

\IEEEoverridecommandlockouts                              

\usepackage{graphicx}

\usepackage{amsmath}
\usepackage{amssymb}
\usepackage{booktabs}
\usepackage{multirow}
\usepackage{diagbox}
\usepackage{graphicx}
\usepackage{subcaption}
\usepackage{caption}
\usepackage{subcaption} 

\usepackage{xcolor} 
\usepackage{array}

\renewcommand{\arraystretch}{1.0}

\usepackage[colorlinks=true,
            linkcolor=blue,
            citecolor=blue,
            urlcolor=blue]{hyperref}

\title{\LARGE \bf
DarkDriving: A Real-World Day and Night Aligned Dataset for Autonomous Driving  in the Dark Environment
}

\author{Wuqi Wang$^{1,*}$, Haochen Yang$^{2,*}$, Baolu Li$^{2}$, Jiaqi Sun$^{1}$, Xiangmo Zhao$^{1}$, Zhigang Xu$^{1}$, Qing Guo$^{3}$, \\ Haigen Min$^{1,\dagger}$, Tianyun Zhang$^{2,\dagger}$, Hongkai Yu$^{2,\dagger}$ 
\thanks{$^{1}$Chang'an University.
$^{2}$Cleveland State University.   
$^{3}$A*STAR.} 
\thanks{$^*$Equal contribution.  $^\dagger$Co-corresponding authors: hgmin@chd.edu.cn, t.zhang85@csuohio.edu, h.yu19@csuohio.edu.}
}

\begin{document}
\maketitle
\input{sec_camera/0_abstract}    
\input{sec_camera/1_intro}

\input{sec_camera/2_relatedwork}

\input{sec_camera/3_dataset}
\input{sec_camera/4_experiment}

\input{sec_camera/5_conclusion}

{
    \small
    \bibliographystyle{IEEEtran}
    \bibliography{main}
}

\end{document}

%% file: sec_camera/0_abstract.tex
\begin{abstract}
The low-light conditions are challenging to the vision-centric perception systems for autonomous driving in the dark environment. In this paper, we propose a new benchmark dataset (named DarkDriving) to investigate the low-light enhancement for autonomous driving. The existing real-world low-light enhancement benchmark datasets can be collected by controlling various exposures only in small-ranges and static scenes. The dark images of the current nighttime driving datasets do not have the precisely aligned daytime counterparts. The extreme difficulty to collect a real-world day and night aligned dataset in the dynamic driving scenes significantly limited the research in this area. With a proposed automatic day-night Trajectory Tracking based Pose Matching (TTPM) method in a large real-world closed driving test field (area: 69 acres), we collected the first real-world day and night aligned dataset for autonomous driving in the dark environment. The DarkDriving dataset has 9,538 day and night image pairs precisely aligned in location and spatial contents, whose alignment error is in just several centimeters. For each pair, we also manually label the object 2D bounding boxes. DarkDriving introduces four perception related tasks, including low-light enhancement, generalized low-light enhancement, and low-light enhancement for 2D detection and 3D detection of autonomous driving in the dark environment. The experimental results show that our DarkDriving dataset provides a comprehensive benchmark for evaluating low-light enhancement for autonomous driving and it can also be generalized to enhance dark images and promote detection in some other low-light driving environment, such as nuScenes.The code and dataset will be publicly available at \url{https://github.com/DriveMindLab/DarkDriving-ICRA-2026}  
\end{abstract}

%% file: sec_camera/1_intro.tex
\section{Introduction}\label{sec:intro}

\begin{figure}[!t]
\centering
\subfloat[DarkDriving Dataset]{%
  \includegraphics[width=.99\columnwidth]{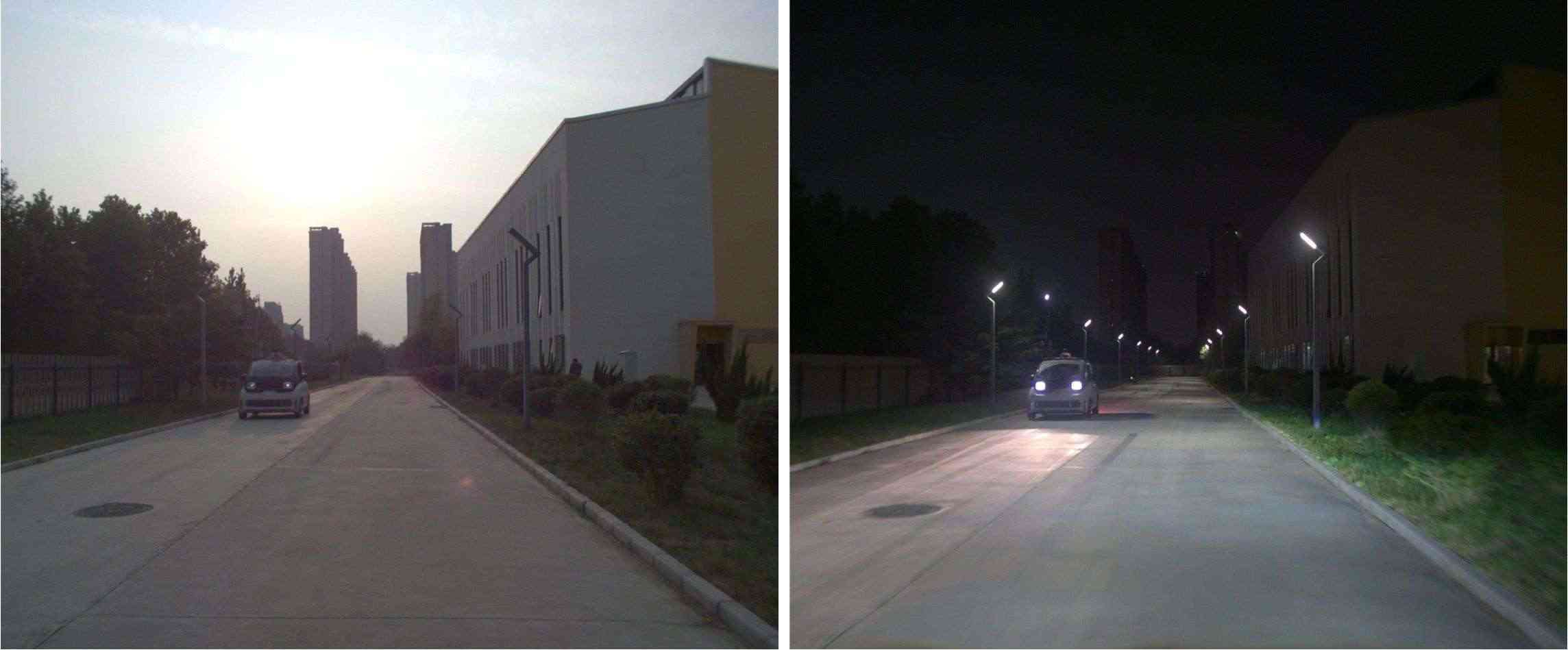}%
}
\hfil
\subfloat[LOL Dataset~\cite{wei2018deep}]{%
  \includegraphics[width=.49\columnwidth,height=.26\columnwidth]{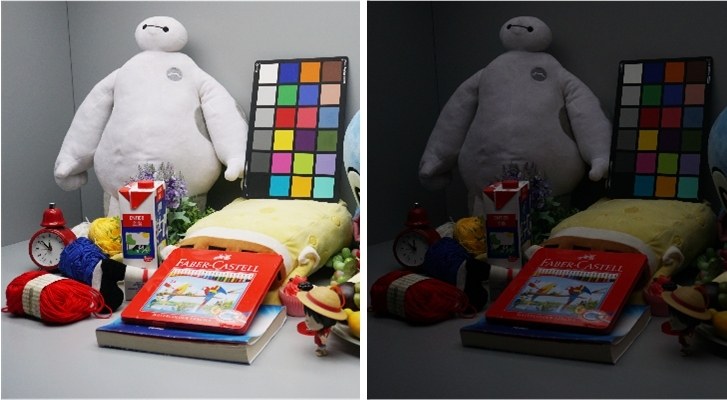}%
}
\hfil
\subfloat[Dark Zurich Dataset~\cite{sakaridis2020map}]{%
  \includegraphics[width=.49\columnwidth,height=.26\columnwidth]{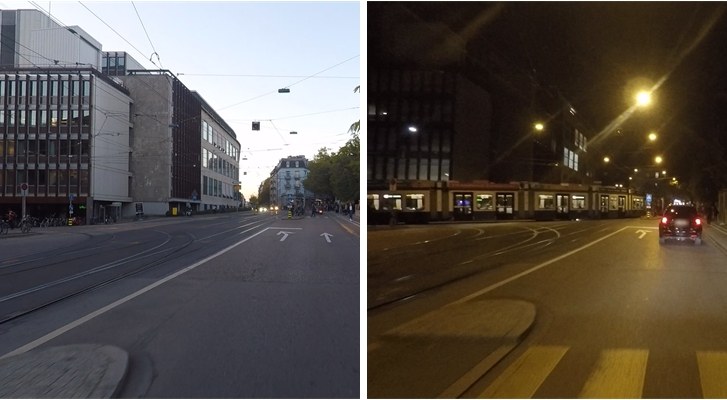}%
}
\vspace{-1mm}
\caption{\textbf{Data examples of our DarkDriving and other representative low-light enhancement and nighttime driving datasets}: (a) our DarkDriving with precisely aligned location and spatial contents (error: centimeters) of day and night, (b) low-light enhancement dataset by controlling exposures, and (c) nighttime driving dataset with roughly aligned daytime location (error: meters) and unaligned spatial contents using the GPS signal.}
\vspace{-2em}
\label{fig:dataset_comparison}
\end{figure}

\begin{table*}[!t]
\setlength{\tabcolsep}{5pt}
\centering
\caption{\textbf{Comparison of our DarkDriving dataset and other low-light enhancement and nighttime driving datasets.}}
\vspace{-0.5em}
\begin{tabular}{l|c|c|c|c|c|c|c}
\hline
\hline
Dataset &
  Year &
  \begin{tabular}[c]{@{}c@{}}Real/\\ Simulated\end{tabular} &
  \begin{tabular}[c]{@{}c@{}}Independent of \\ Exposure Control\end{tabular} &
  \begin{tabular}[c]{@{}c@{}}RGB\\ Images\end{tabular} &
  Scene &
  \begin{tabular}[c]{@{}c@{}} Day and Night\\ Paired Images\end{tabular} &
  \begin{tabular}[c]{@{}c@{}} Pair Alignment \\ Error\end{tabular} \\ \hline
LOL~\cite{wei2018deep} &
  2018 &
  Real &
  $\times$&
  1,000 &
  Static &
  \checkmark &
  - \\
SCIE~\cite{cai2018learning} &
  2018 &
  Real &
  $\times$&
  4,413 &
  Static &
  \checkmark &
  - \\
SID~\cite{chen2018learning} &
  2018 &
  Real &
  $\times$&
  5,094 &
  Static &
  \checkmark &
  - \\ \hline
BBD100K-dark~\cite{yu2020bdd100k} &
  2020 &
  Real &
  \checkmark&
  14K &
  Driving &
  $\times$&
  - \\
NuScenes-night~\cite{caesar2020nuscenes} &
  2020 &
  Real &
  \checkmark&
  4.6K &
  Driving &
  $\times$ &
  - \\
Dark Zurich~\cite{sakaridis2020map} &
  2020 &
  Real &
  \checkmark&
  8,779 &
  Driving &
  \checkmark &
  Meters \\ 
NightCity~\cite{tan2021night} &
  2021 &
  Real &
  \checkmark&
  4,297 &
  Driving &
  $\times$ &
  - \\
SHIFT~\cite{sun2022shift} &
  2022 &
  Simulated &
  \checkmark &
  2.5M &
  Driving &
  $\times$ &
  - \\ \hline
\textbf{DarkDriving (Ours)} &
  \textbf{2026} &
  \textbf{Real} &
  \textbf{\checkmark} &
  \textbf{19,076} &
  \textbf{Driving} &
  \textbf{\checkmark} &
  \textbf{Centimeters} \\ \hline \hline
\end{tabular}
\label{tab:datasets}
\vspace{-1em}
\end{table*}

The vision-centric perception systems are widely used on the intelligent vehicles due to the effective cost of camera sensors compared to the expensive cost of LiDAR sensors, such as the Tesla's Full Self-Driving (FSD) system. The vision-centric perception systems are reliable to perceive the nearby environment in most of driving scenarios, however the low-light conditions in the dark environment might result in the significant challenges to the camera based vision-perception systems. For example, the nighttime fatal rate is 18\% higher than the daytime fatal rate~\cite{ashraf2019catastrophic}, and an Uber self-driving vehicle failed to detect a 
pedestrian and killed that pedestrian in Arizona at one dark night of 2018~\cite{ntsb_uber_2018}. These statistical data and driving fatalities during nighttime increase the demands to investigate autonomous driving in the dark environment.

One recent study~\cite{li2024light} has demonstrated that advanced low-light enhancement methods can improve the robustness of vision-centric perception systems for autonomous driving at night. However, there is currently no real-world dataset providing day and night paired data specifically for investigating low-light enhancement in autonomous driving. Existing real-world low-light enhancement datasets typically acquire paired images either by varying camera exposures or by capturing static scenes under different lighting conditions within small regions, \textit{e.g.}, the LOL dataset~\cite{wei2018deep}.  Nighttime driving datasets can provide approximate daytime counterparts using GPS signals (error: meters), but they fail to ensure spatial content alignment due to moving objects and pose variations, \textit{e.g.}, the Dark Zurich dataset~\cite{sakaridis2020map}. Due to these limitations, some studies resort to synthetic datasets by artificially darkening daytime images~\cite{li2024light}, yet such methods do not realistically simulate dynamic nighttime conditions, including vehicle and street lighting variations. The lack of a real-world dataset with precisely aligned day and night image pairs presents a significant challenge in this field.

This paper introduces a real-world benchmark dataset, called DarkDriving, to advance research on low-light enhancement for autonomous driving. Leveraging the large-scale closed driving test field (69 acres) at Chang'an University, we develop a novel Trajectory Tracking based Pose Matching (TTPM) method to automatically collect day and night image pairs under real-world driving conditions. The TTPM method ensures that the autonomous vehicle precisely follows the same trajectory at night as it did during the day, achieving fine-grained alignment in both location and spatial content, with an alignment error within centimeters. The proposed DarkDriving dataset consists of 9,538 precisely aligned day and night image pairs, with manually annotated objects. A comparative analysis with other representative datasets is presented in Fig.~\ref{fig:dataset_comparison} and Table~\ref{tab:datasets}. To further explore autonomous driving in low-light conditions, we extend our study to include four perception tasks for autonomous driving in dark environments, namely: low light enhancement, generalized low-light enhancement, and low-light enhancement for 2D detection and 3D detection. Our contributions are summarized as follows.

\begin{itemize}
    \item We propose the first real-world benchmark dataset (DarkDriving) for studying low-light enhancement in autonomous driving, containing 9,538 day and night aligned image pairs collected from a large-scale closed driving test field (69 acres). 

    \item We design a novel automatic Trajectory Tracking based Pose Matching (TTPM) method, which ensures precise location and spatial content alignment for each day and night image pair with an alignment error of just a few centimeters. 

    \item Using the proposed DarkDriving dataset, we investigate four perception tasks, \textit{i.e.}, low-light enhancement, generalized low-light enhancement, and low-light enhancement for 2D detection and 3D detection, to advance research on nighttime  autonomous driving.
\end{itemize}

%% file: sec_camera/2_relatedwork.tex
\section{Related Work} \label{sec:relatedwork}
\subsection{Datasets for Low-light Enhancement}

The task of low-light image enhancement focuses on recovering obscured details in poorly illuminated regions, thereby improving the image visibility. In supervised learning approaches, paired datasets play a crucial role. Several publicly available datasets provide such paired images for low-light enhancement~\cite{afifi2021learning, bychkovsky2011learning, wang2019underexposed, wei2018deep, cai2018learning, chen2018learning}. The LOL dataset~\cite{wei2018deep} was the first to introduce real-scene low/normal-light image pairs for enhancement tasks. SCIE~\cite{cai2018learning} and SID~\cite{chen2018learning} offer images captured at different exposure levels.  However, these datasets only cover small static scenes not designed for autonomous driving, so they are not suitable for investigating the dark driving problem. 
\vspace{-0.3em}


\subsection{Datasets for Nighttime Driving}
\vspace{-0.2em}

Several datasets have been developed to support nighttime autonomous driving research. NuScenes-night~\cite{caesar2020nuscenes} was the first dataset collected from an autonomous vehicle operating on public roads at night, including LiDAR, cameras, and radar data. BDD100K~\cite{yu2020bdd100k} contains a subset of nighttime driving scenes (BDD100K-dark) to benchmark image recognition algorithms in low-light conditions. Dark Zurich~\cite{sakaridis2020map} offers a collection of images captured at different times, including roughly matched day-and-night image pairs aligned using GPS data. However, because of the coarse alignment and the lack of precisely paired images, these datasets are not ideal for supervised low-light enhancement training. Although simulation-based approaches~\cite{sun2022shift} offer the advantage of generating paired day-night images, they cannot accurately capture the complexities of real-world 
nighttime driving. Despite these efforts, no existing dataset provides precisely aligned real-world day-and-night images for autonomous driving in dark.

%% file: sec_camera/3_dataset.tex
\section{DarkDriving Dataset}\label{sec:data}

\begin{figure}[!t]
   \centering
   \includegraphics[width=1\columnwidth]{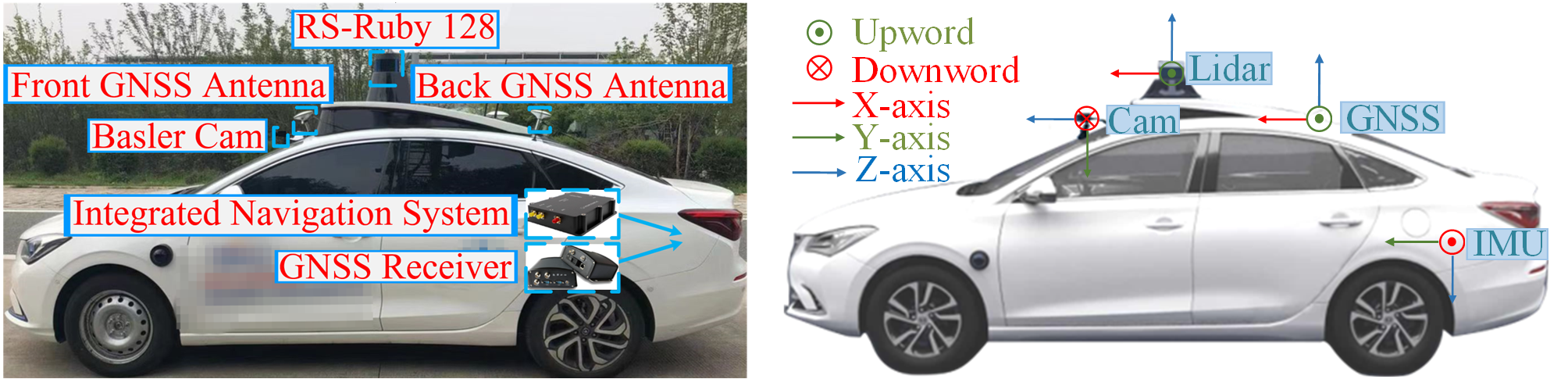}
   \vspace{-1.5em}
   \caption{\textbf{Setup of sensors on the automated vehicle and their coordinate systems.}}
   \label{fig:CAV} 
\end{figure}

\begin{table}[!t]
\caption{\textbf{Specification of sensors on automated vehicle.}}\label{tab:sensor_specs}
\vspace{-0.5em}
\setlength{\tabcolsep}{5pt}
\centering
\renewcommand{\arraystretch}{1.2} 
\begin{tabular}{l|p{0.7\columnwidth}}
\hline
\textbf{Sensors} & 
\textbf{Details} 
\\ \hline
1×Camera & 
RGB, 2,448×2,048 resolution, 41° FOV 
\\ \hline
1×LiDAR& 
\begin{tabular}{@{}p{0.7\columnwidth}}
128 channels,  200m capturing  range,   
-25° to +15° vertical FOV, Range  
Precision: ±2cm error\end{tabular}
\\ \hline
GPS \& IMU & 
Xsens MTi-680G, 400 Hz SDI    
\\ \hline
\end{tabular} 
\end{table}

\begin{figure}[htbp]
   \centering
   \includegraphics[width=1\columnwidth]{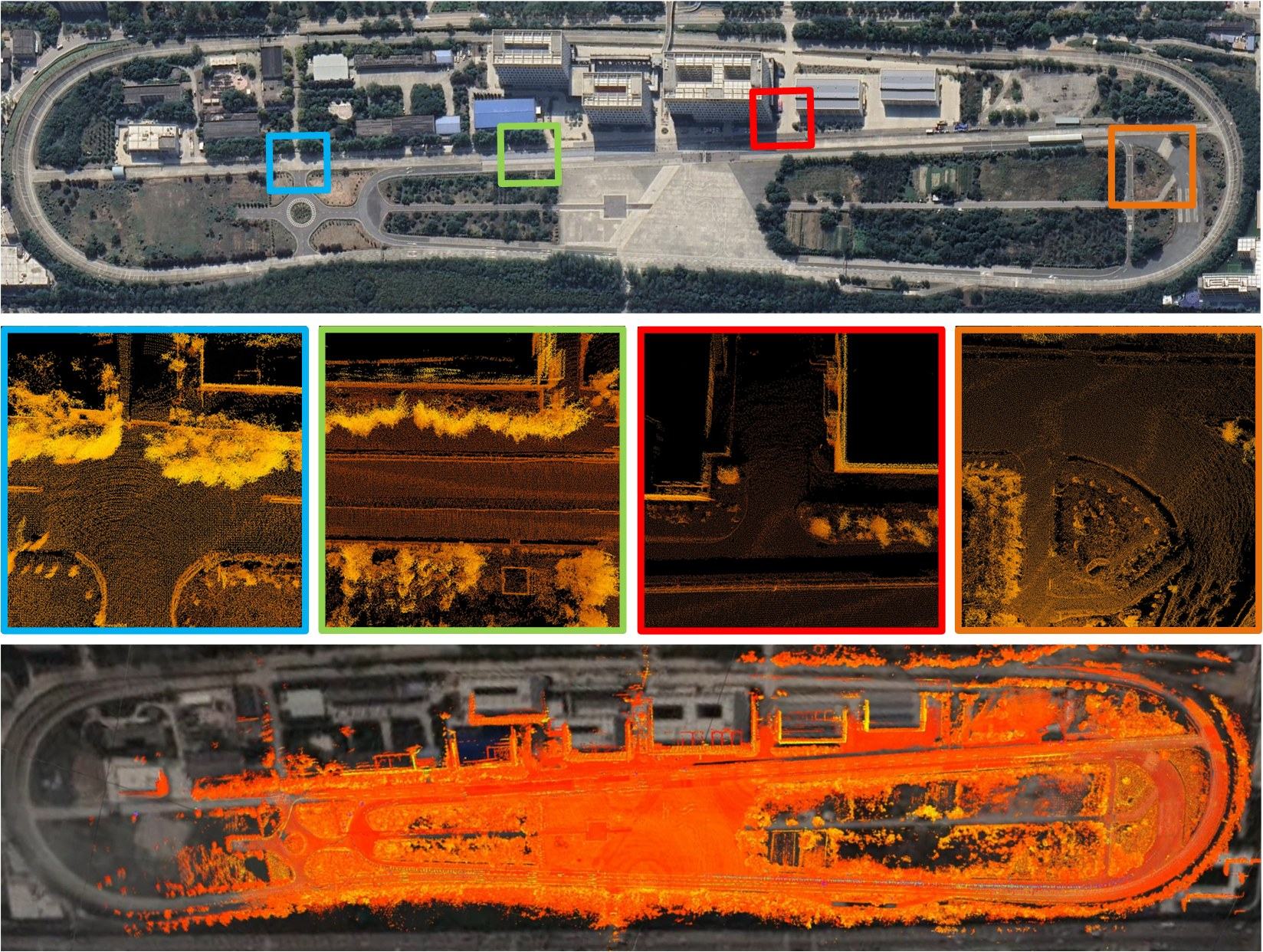}
   \vspace{-1.5em}
   \caption{\textbf{Details of the large real-world closed driving test field at Chang'an University.} From top to bottom:  satellite image of driving test field, local and global demos of its High-Precision Point Cloud Map.}
   \label{fig:test_field}
   \vspace{-1.5em}
\end{figure}

\subsection{Automated Vehicle and Sensor Setup} 

We collect the DarkDriving dataset using one automated vehicle and multiple background vehicles. The Fig.~\ref{fig:CAV} shows the automated vehicle (moving) used for our data collection. The background vehicles (static in the defined locations during day and night) consist of several normal vehicles, which are employed to simulate diverse urban traffic scenarios. The automated vehicle is equipped with 128-channel LiDAR, an industrial front-facing camera, and an integrated navigation system. The layout of the automated vehicle sensors is also illustrated in Fig.~\ref{fig:CAV}, and the detailed sensor specification is provided in Table~\ref{tab:sensor_specs}. Theses sensor measurements use the same integrated navigation system for time synchronization, ensuring that the point cloud, camera data, and integrated navigation data within the same sequence are time-synchronized at the hardware level.

\subsection{Large Real-world Closed Driving Test Field} 

\begin{figure*}[htbp]
   \centering
   \includegraphics[width=0.85\textwidth]{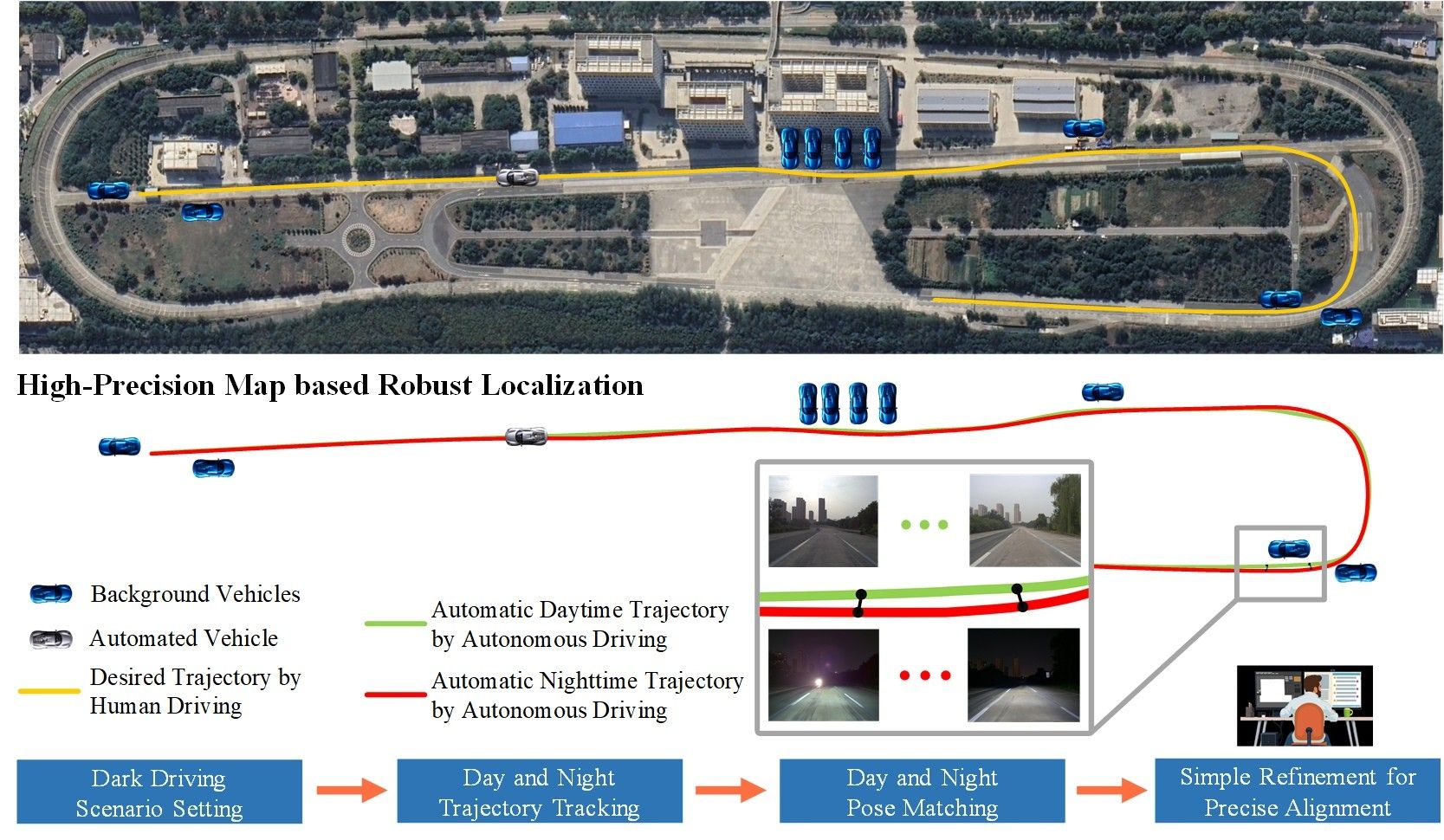}
   \vspace{-2.5mm}
   \caption{\textbf{The pipeline of the automatic day-and-night Trajectory Tracking based Pose Matching (TTPM) method for data collection.} One example of the desired trajectories is shown here. Best view in color.}
   \vspace{-2em}
   \label{fig:Framework}
\end{figure*}

To collect the day and night data of the consistent dynamic driving scenarios, an automated vehicle is used to collect the data in a large real-world closed driving test field. There are several  advantages to use this driving test field for our DarkDriving data collection. The driving test field spans a large-scale  area of 69 acres and includes over 40 typical vehicle driving scenarios in the real world, such as multi-lane road, curved road, intersection, streetlight and so on, which could well represents the real-world driving scenes in the dark environment. Because this driving test field is closed, we could avoid the inconsistency of moving objects in day and night and also pre-define and control the location and lighting conditions of the static background vehicles and streetlights to represent diverse  nighttime driving scenarios. This driving test field provides us a High-Precision Point Cloud Map (constructed based on the LIO-SAM method~\cite{shan2020lio} before) that built a digital-twin version of the real-world driving test field, whose mapping error is in just centimeters. Fig.~\ref{fig:test_field} shows the satellite image of the test field and the local, global demonstrations of its High-Precision Point Cloud Map.


\subsection{Data Acquisition by TTPM} 
The pipeline of our TTPM method is shown in Fig.~\ref{fig:Framework}. Let us introduce the details of each module in this section. 



\subsubsection{High-Precision Map based Robust Localization}

Since GNSS is prone to the multipath effects in some complex driving environments, DarkDriving uses the available High-Precision Point Cloud Map (defined as $\mathbf{M}$) of the real-world closed driving test field for robust vehicle pose estimation. We provide continuous and accurate localization by associating the LiDAR-based point cloud data with the High-Precision Map using the Normal Distributions Transform (NDT)~\cite{biber2003normal}. First of all, we uniformly divide the High-Precision Point Cloud Map $\mathbf{M}$ into local grids, where each local grid can be represented as a Gaussian distribution $\mathbf{G}_i$. For each frame, the total $m$ points   $\mathbf{c} \in \mathbb{R}^{4 \times m}$ of the automated vehicle can be obtained via its LiDAR sensor, where each point is a 4D coordinate vector $[X, Y, Z, I]$. Then, NDT fits $\mathbf{c}$ into the different local  Gaussian distributions $\{\mathbf{G}_i \mid i= 1,2,...,k\}$ via a Homogeneous Transformation Matrix (HTM) $\mathbf{h} \in \mathbb{R}^{4 \times 4}$ to find the Mahalanobis Distance based association error:  

\begin{equation}
\begin{aligned}
    \mathbf{E(\mathbf{h})} = \sum_{i=1}^{k} (\mathbf{h}\mathbf{c} - \mathbf{\mu_i})^\text{T} \mathbf{\Sigma_i}^{-1} (\mathbf{h}\mathbf{c} - \mathbf{\mu_i}),
\end{aligned}
\end{equation}
where  $\mathbf{\mu_i}$ and $\mathbf{\Sigma_i}$ are the mean and covariance of the $\mathbf{G}_i$ in the map $\mathbf{M}$. The objective is to minimize the association error to determine the optimal HTM for each frame: 

\begin{equation}
\begin{aligned}
    \mathbf{h^*} = \arg\min_{\mathbf{h}} \mathbf{E}(\mathbf{h}).
\end{aligned}
\end{equation}

After obtaining the optimal HTM $\mathbf{h^*}$ for one frame, using  $\mathbf{M}$ as the coordinate system, the robust localization  (vehicle pose) can be estimated by the homogeneous transformation. Finally, for that frame, let us re-define the estimated vehicle pose as $\mathbf{p}=[x, y, z, roll, yaw, pitch]$ after the optimal homogeneous transformation.

\subsubsection{Day and Night Trajectory Tracking}

To ensure the precisely aligned location and spatial contents of captured images during day and night, it is important to maintain the consistent moving trajectory (a time sequence of pose and velocity) of the automated vehicle during day and night. Because human driver cannot guarantee a perfectly uniform trajectory across multiple trips, substantial discrepancies in yaw and position are inevitable. Therefore, the automatic Trajectory Tracking method, widely used in robot control, is used to keep consistent trajectory of the automated vehicle during day and night.

First, we define a driving scenario in the closed driving test field and manually drive the automated vehicle for that driving scenario, where the vehicle's pose and velocity are recorded, leading to a desired driving trajectory $\mathbf{T}$. To facilitate the later processing and avoid motion blur, the desired driving trajectory $\mathbf{T}$ has a rough velocity of 9 miles/hour in the manual human driving. Then, we make the automated vehicle autonomously drive and follow  $\mathbf{T}$  during daytime, resulting in an automatic daytime trajectory $\mathbf{T^d}$. Furthermore, we let the automated vehicle autonomously drive and follow  $\mathbf{T}$ during nighttime, generating an automatic nighttime trajectory $\mathbf{T^n}$. 

\vspace{-0.2em}

Our Trajectory Tracking during day and night is realized by the Pure Pursuit algorithm~\cite{macenski2023regulated}. The Pure Pursuit algorithm $\mathbf{\Phi}$ provides the motion planning for the next time steps, by computing the minimum Euclidean distance based on the vehicle's current pose, velocity, and the desired trajectory $\mathbf{T}$, which can be represented as:

\begin{equation}
[\mathbf{p}_{t+\Delta t}, \mathbf{v}_{t+\Delta t}] = \mathbf{\Phi}(\mathbf{p}_t, \mathbf{v}_t, \mathbf{T}),  
\end{equation}
where $\mathbf{\Phi}$ takes the current pose $\mathbf{p}_t$, current velocity $\mathbf{v}_t$ at the time step $t$  and the desired trajectory $\mathbf{T}$ as the input and outputs the pose and velocity $[\mathbf{p}_{t+\Delta t}, \mathbf{v}_{t+\Delta t}]$ for motion planning in the future $\Delta t$ time steps. These motion planning outputs are then fed to the PID (Proportional Integral Derivative) Control module to obtain the vehicle acceleration and steering angle for the automatic vehicle control. With this day and night Trajectory Tracking method, we could keep the daytime trajectory $\mathbf{T^d}$ and nighttime trajectory $\mathbf{T^n}$ consistent for the automated vehicle.

\subsubsection{Day and Night Pose Matching}
After day and night trajectory tracking, we get 23 day and night trajectory pairs and we denote each day and night trajectory pair as $[\mathbf{T^d}, \mathbf{T^n}]$. There are three  challenges to find precisely aligned day and night image pairs. The trajectories $\mathbf{T^d}$ and $\mathbf{T^n}$ are highly consistent, but they might still have small control inconsistency due to the slight differences of vehicle weight and roadway surface in day and night. In our data collection, the collection time of different sensors is synchronized with same starting time, but they have different frequencies. The LiDAR based pose collection rate is high for robust localization (fps: 20), and the camera image collection rate during daytime is low for clear image details (fps: 10), while the camera image collection rate during nighttime is lower for more camera exposure time (fps: 6). Directly matching  images in day and night is challenging because of significantly different image and lighting features in day and night.

To solve the matching challenges, we design a Day and Night Pose Matching method to find the precisely aligned day and night image pairs. Given a query camera image in daytime (camera image frame index: $q$), we can easily find its corresponding pose index in daytime by the nearest neighbor search in the time domain, denoted as a mapping function $\psi(\cdot)$, leading to the corresponding daytime pose $\mathbf{p}^{d}_{\psi(q)}$. Then we search for the minimum Euclidean distance over all the nighttime poses $\{ \mathbf{p}^{n}_{\psi(f)} \}$ by solving the following optimization problem:

\begin{equation}
\begin{aligned}
\underset{f}{\text{Minimize }} & \| \mathbf{p}^{d}_{\psi(q)} - \mathbf{p}^{n}_{\psi(f)} \|_2, \\
\text{s.t. } & \| \mathbf{p}^{d}_{\psi(q)} - \mathbf{p}^{n}_{\psi(f)} \|_2 \leq \delta \\
& 1 \leq f \leq F  
\end{aligned}
\end{equation}
where $f$ is the camera image frame index and $F$ is the total image frame number in the nighttime trajectory $\mathbf{T^n}$, and $\delta$  represents the maximum acceptable pose matching error, which is set to 5cm in this paper. In this way, after this pose matching based optimization, the day and night image pairs can be precisely aligned  automatically.

\begin{figure*}[htbp] 
\centering
\begin{subfigure}[t]{0.333\textwidth} 
   \centering
   \includegraphics[width=\linewidth]{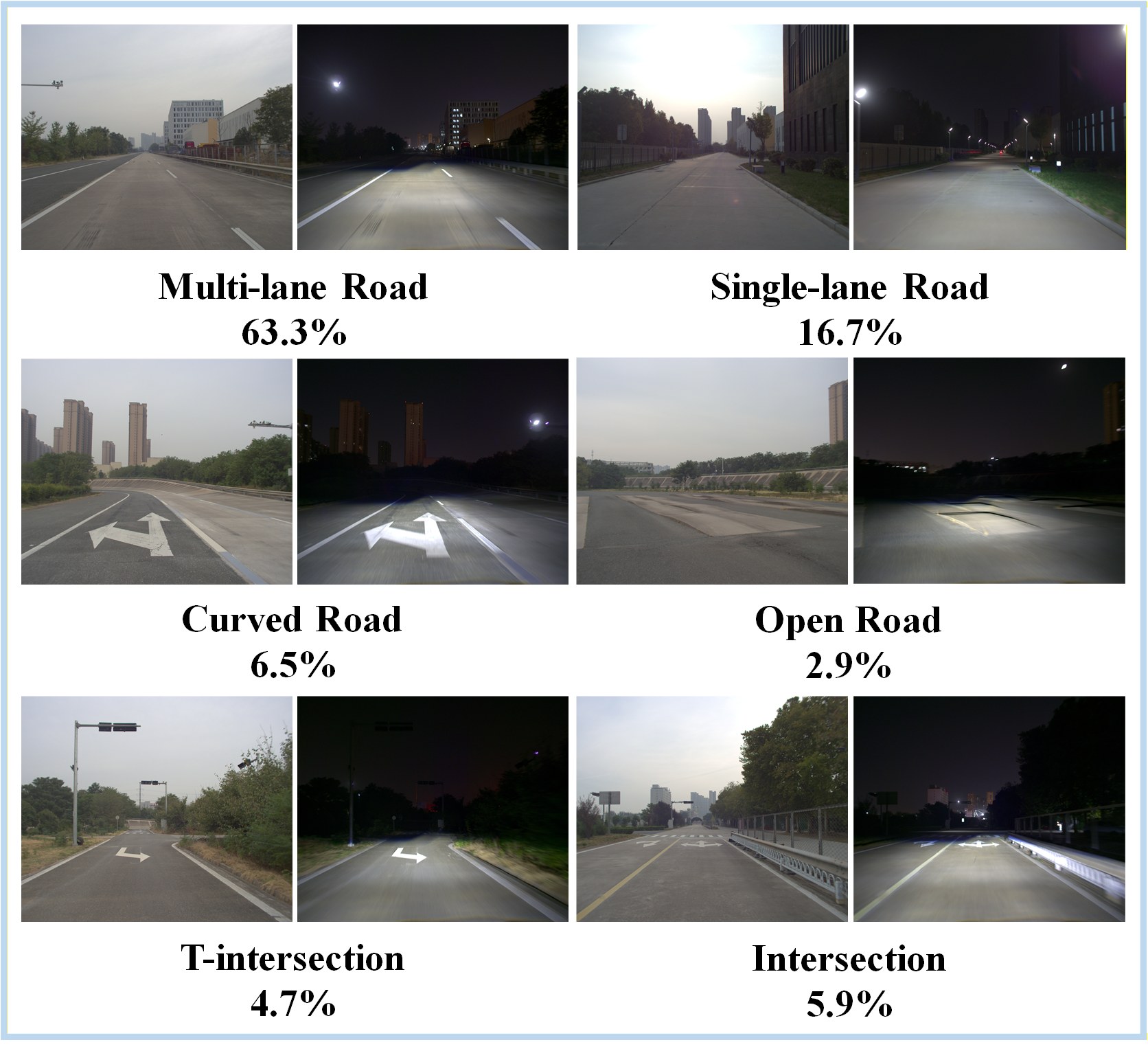}
   \vspace{-1.2em}
   \caption{Diverse Types of Road Scenes}
\end{subfigure}
\hfill 
\begin{subfigure}[t]{0.66\textwidth} 
   \centering
   \includegraphics[width=\linewidth]{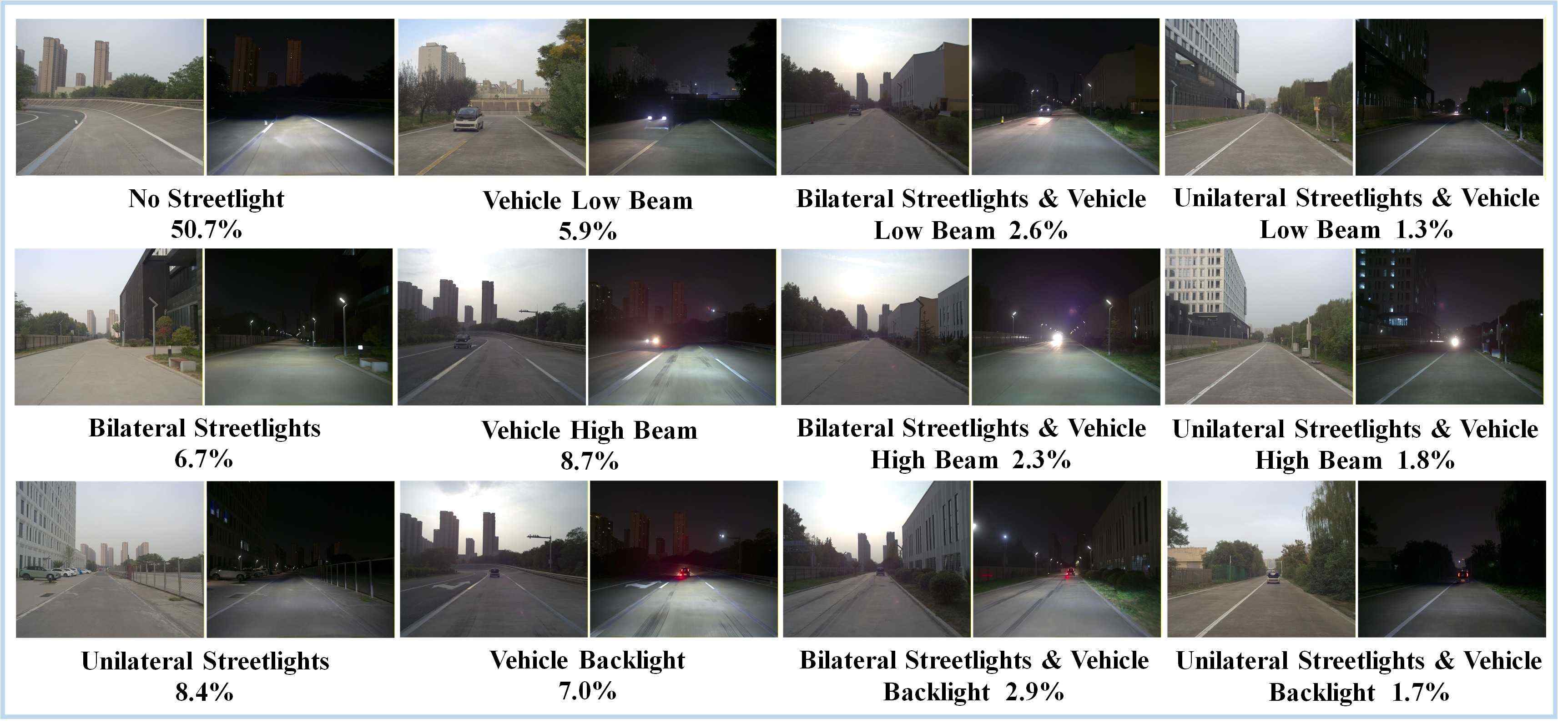}
   \vspace{-1.2em}
   \caption{Various Nighttime Lighting Conditions}
   \label{fig:light}
\end{subfigure}%
\vspace{-0.3em}
\caption{\textbf{Dark driving scenarios of our DarkDriving dataset.} One sample day-night image pair is shown for each dark driving scenario, followed by its percentage distribution in our dataset.}
\vspace{-1.5em}
\label{fig:scenarios}
\end{figure*}

\subsubsection{Simple Refinement for Precise Alignment}
Our data collection took 8 days in the large real-world driving test field, and the automatic Day and Night Trajectory Tracking generated 23 pairs of 
day-night image sequences, which was a total of over 60K images. To increase the data diversity, among them, 12 paired sequences have different desired driving trajectories $\mathbf{T}$. For the remaining paired sequences with shared $\mathbf{T}$, we set different locations and lighting conditions of background vehicles. The Day and Night Pose Matching automatically output 13,811 day-night image pairs with aligned location and spatial contents.

However, during our data collection, sometimes uncontrollable traffic participants might occasionally appeared in the camera. For example, uncontrollable staffs and third-party testing vehicles of the large driving test field might be occasionally captured by the camera in the daytime driving, while they were not recorded in the nighttime driving. In addition, the accumulated errors in the engineering implementation might accidentally result in the decimeter-level day and night alignment error. To further ensure the accuracy of the day and night correspondence, 4 human drivers were recruited, and each spent 5 hours to review the collected data to remove the day and night image pairs with different dynamic objects and the decimeter-level alignment errors. After this simple human refinement, we finally obtained 9,538 day and night image pairs with highly precise alignment of location and spatial contents, whose correspondence error is just in centimeter-level. The temporal consistency is significantly broken for the collected time-series image sequence after the Day and Night Pose Matching and the Simple Refinement, so we deal with each day-night image pair independently.     


\subsection{Dark Driving Scenario Setting and Analysis} 
To represent real nighttime driving scenarios, we collected data on 6 typical road scenarios in the large real-world driving test field: Multi-lane Road, Single-lane Road, Curved Road, Open Road, T-intersection, Intersection. By setting  streetlights and beam conditions of background vehicles (static in the defined locations during day and night), we set 12 lighting conditions in nighttime driving: No Streetlight, Vehicle Low Beam, Bilateral Streetlight \& Vehicle Low Beam,  Unilateral Streetlight \& Vehicle Low Beam, Bilateral Streetlight, Vehicle High Beam,  Bilateral Streetlight \& Vehicle High Beam,  Unilateral Streetlight \& Vehicle High Beam,  Unilateral Streetlight, Vehicle Backlight, Bilateral Streetlight \& Vehicle Backlight, Unilateral Streetlight \& Vehicle Backlight.

We think setting diverse road scenarios and different lighting conditions of streetlights and background vehicles is necessary to describe the various traffic scenes in the real-world nighttime driving, so we define the above multiple dark driving scenarios in this research. The Fig.~\ref{fig:scenarios} shows one sample day-night image pair and the percentage distribution of each dark driving scenario in our DarkDriving.

\subsection{Data Processing and Annotation} 

Our dataset contains 9,538 pairs of precisely aligned day-night RGB images, where the  image resolution is 2,448×2,048. Then, we randomly split each day-night sequence pair, which generates a training set of 5,906 image pairs and a testing set of 3,632 image pairs to train and test deep learning models for autonomous driving perception in the dark environment. The precisely aligned image pair data in our  DarkDriving is excellent to train the deep learning models, especially for supervised learning.     


For the task of Low-light Enhancement for 2D detection, we manually annotated the object 2D bounding boxes in each image of our DarkDriving dataset, leading to 13,184 annotated 2D  bounding boxes. Because the data was collected in a closed driving test field without many traffic participants, we only labeled the major object class (Car) in the annotation.



%% file: sec_camera/4_experiment.tex
\section{Experiments}
\label{sec:exp}

In our experiments, four autonomous driving related perception tasks are investigated.

\definecolor{red}{rgb}{1,0,0}
\definecolor{blue}{rgb}{0,0,1}

\begin{table*}[!t]
    \centering
    \caption{\textbf{Quantitative comparison of image quality for SOTA low-light enhancement methods on the testing set of DarkDriving}. The best and second best results are highlighted in \textbf{bold} and \underline{underline}, respectively.}
    \resizebox{0.8\textwidth}{!}{
    \begin{tabular}{l|c|c|c|c|c|c|c}
        \toprule
        & \multicolumn{3}{c|}{Full-reference Metrics} & \multicolumn{4}{c}{Non-reference Metrics} \\ 
        \cmidrule(lr){2-4} \cmidrule(lr){5-8} 
        {Methods} & {PSNR↑} & {SSIM↑} & {LPIPS↓} & {MUSIQ↑} & {NIQE↓} & {HyperIQA↑} & {CNNIQA↑}\\ 
        \midrule
        Night Image           
        & 6.58 & 0.38 & 0.59 
        & 47.79 & 6.58 & 0.34 & \underline{0.54} \\ 
        \midrule
        LLFormer       
        & 22.95 & 0.79 & 0.41 
        & 26.81 & 6.28 & 0.25 & 0.27 \\

        SNR-SKF          
        & 20.29 & 0.74 & 0.39 
        & 35.80 & 5.86 & 0.32 & 0.40 \\

        Retinexformer       
        & \underline{23.66} & \underline{0.80} & \underline{0.34} 
        & 33.18 & 6.06 & 0.31 & 0.34 \\

        ControlNet 
        & 13.17 & 0.58 & 0.50 
        & \textbf{65.16} & \underline{5.32} & \textbf{0.66} & \textbf{0.67} \\

        SNR-Aware   
        & \textbf{28.55} & \textbf{0.81} & \textbf{0.15} 
        & \underline{57.80} & \textbf{4.44} & \underline{0.51} & 0.51 \\
        \bottomrule
    \end{tabular}
    }
    \label{tab:low-light-enhancement}
\end{table*}

\begin{figure*}[t]
\centering
\setlength{\tabcolsep}{0.5pt} 
\renewcommand{\arraystretch}{0.95} 

\newcommand{\CellW}{.141\textwidth}
\newcommand{\CellH}{.141\textwidth}

\begin{tabular}{@{}ccccccc@{}}
\includegraphics[width=\CellW,height=\CellH]{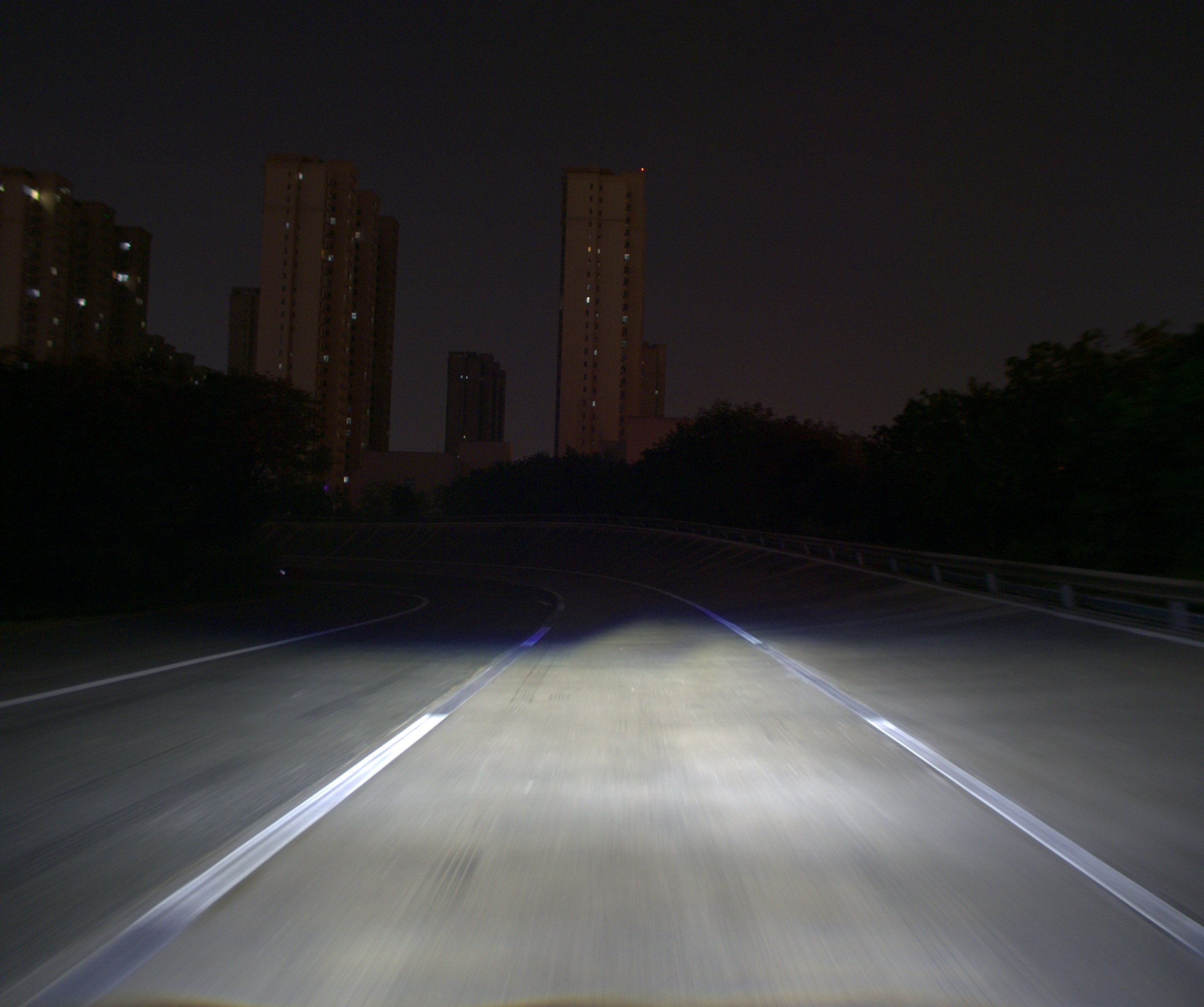} &
\includegraphics[width=\CellW,height=\CellH]{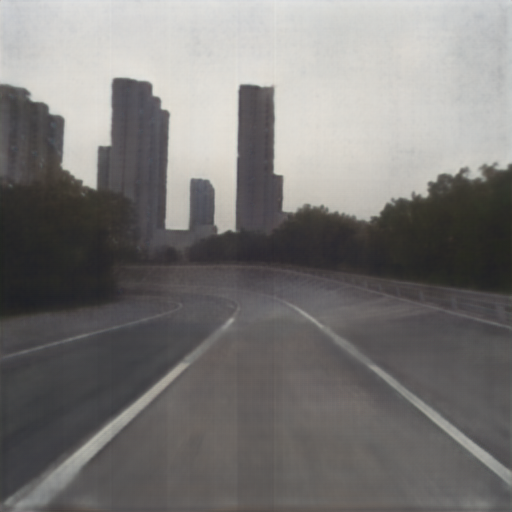} &
\includegraphics[width=\CellW,height=\CellH]{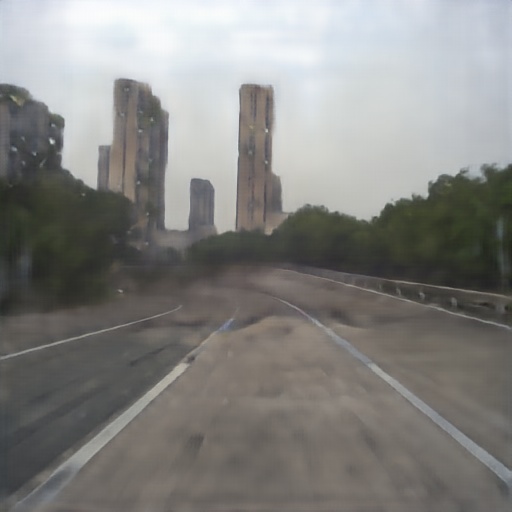} &
\includegraphics[width=\CellW,height=\CellH]{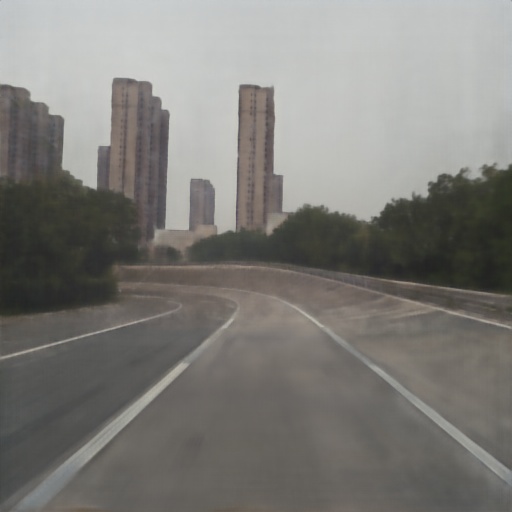} &
\includegraphics[width=\CellW,height=\CellH]{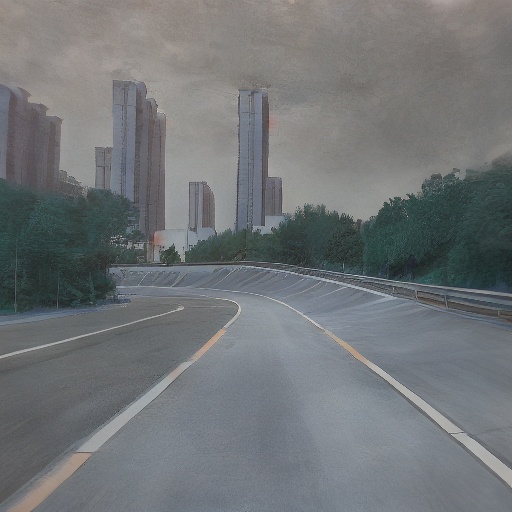} &
\includegraphics[width=\CellW,height=\CellH]{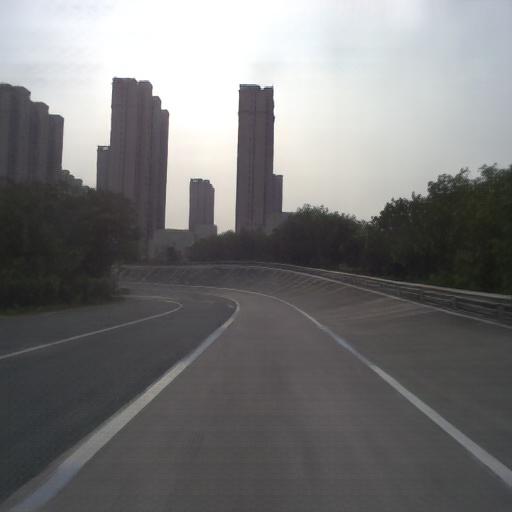} &
\includegraphics[width=\CellW,height=\CellH]{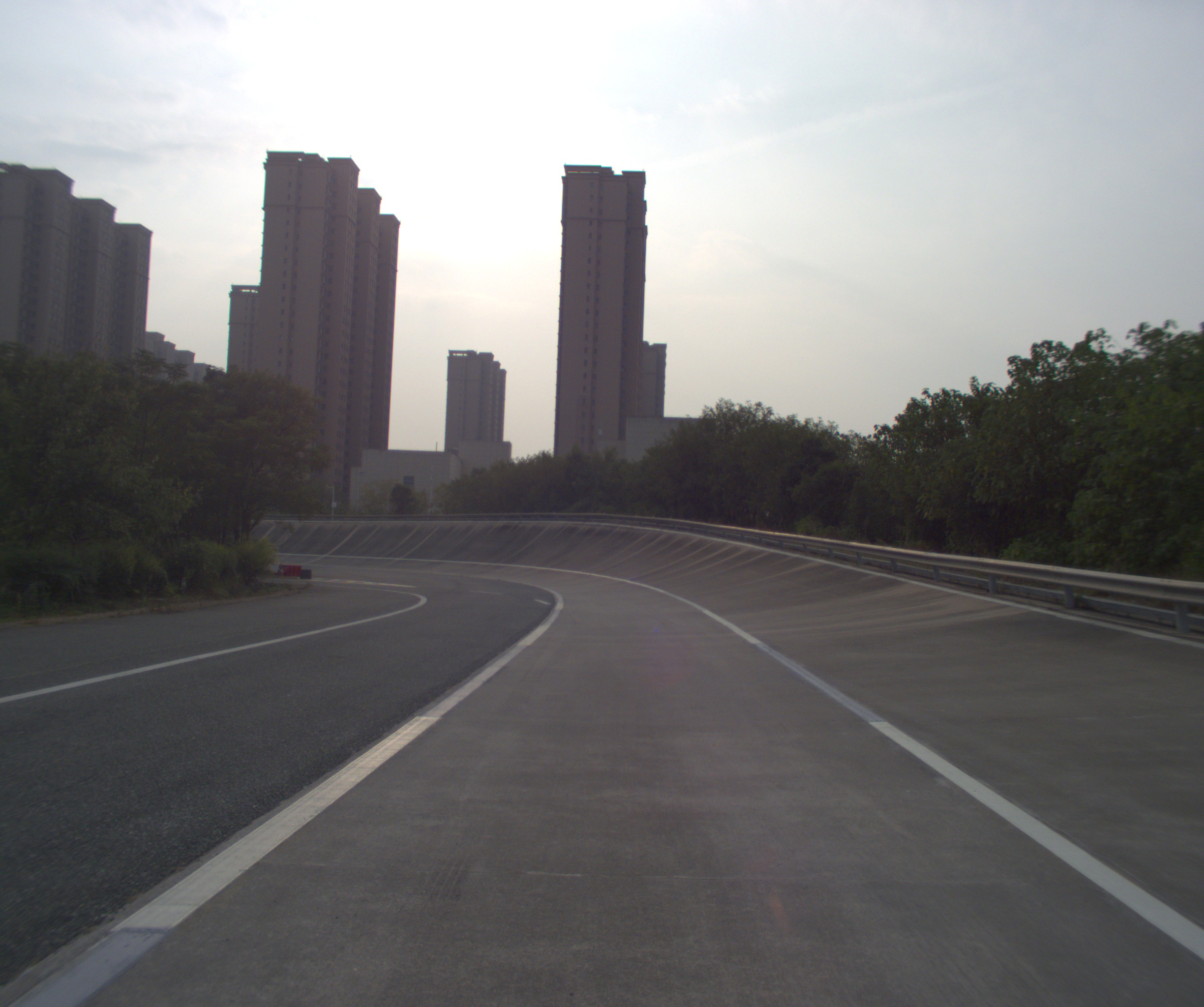} \\[-2pt]

\includegraphics[width=\CellW,height=\CellH]{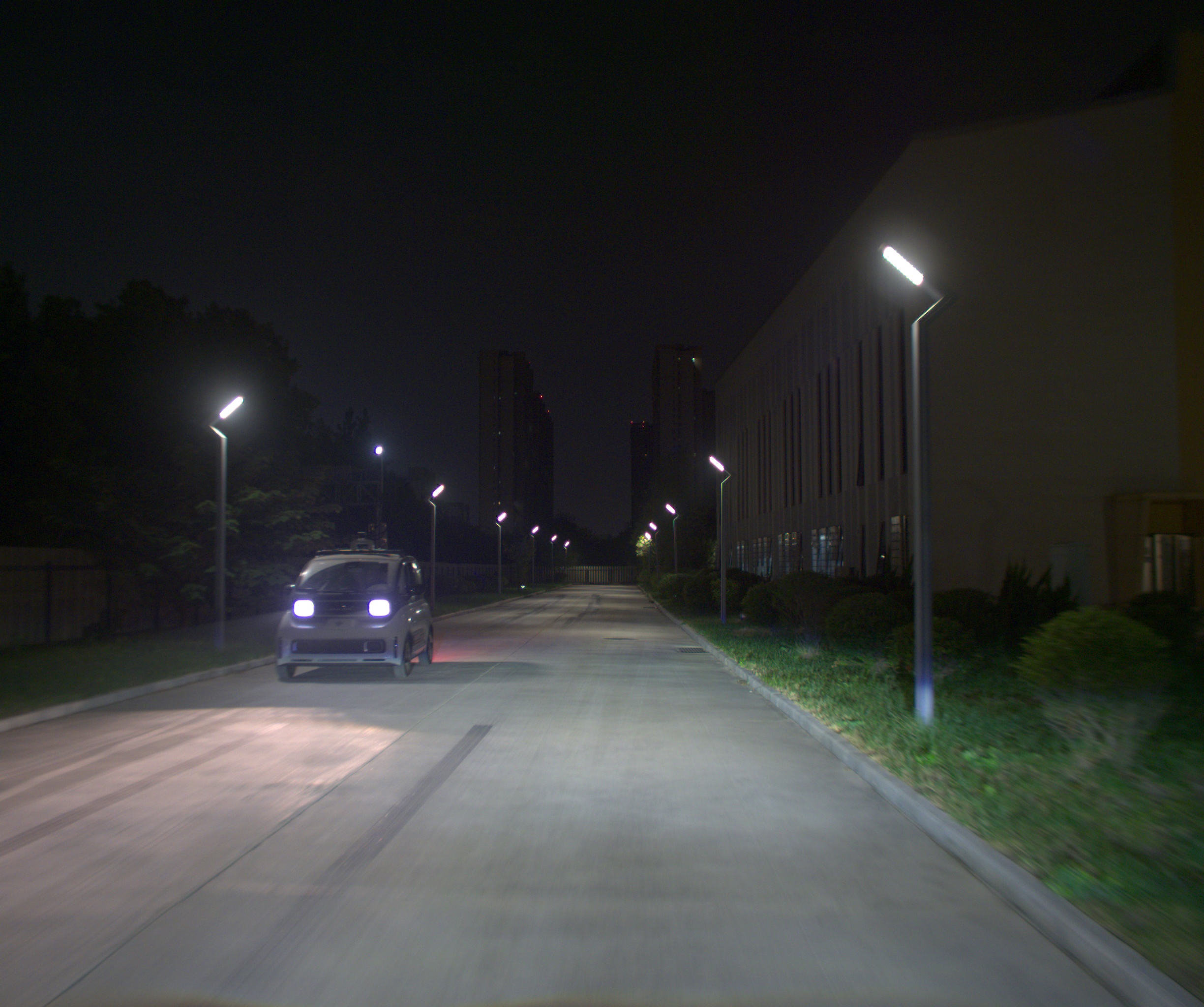} &
\includegraphics[width=\CellW,height=\CellH]{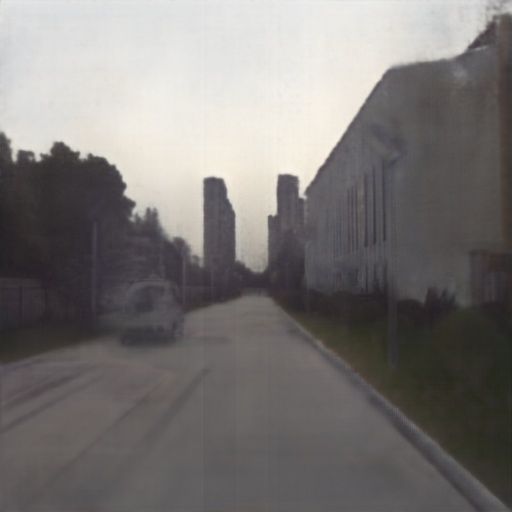} &
\includegraphics[width=\CellW,height=\CellH]{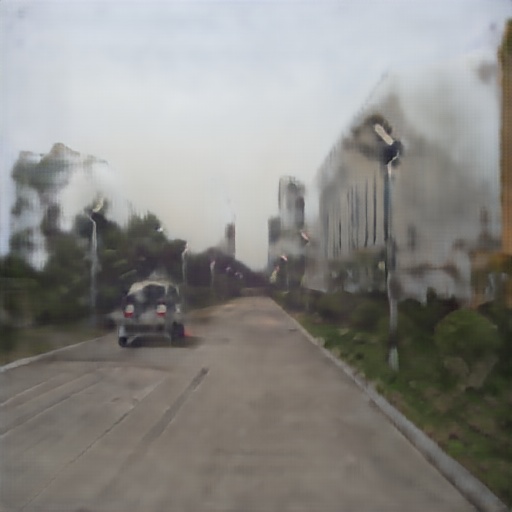} &
\includegraphics[width=\CellW,height=\CellH]{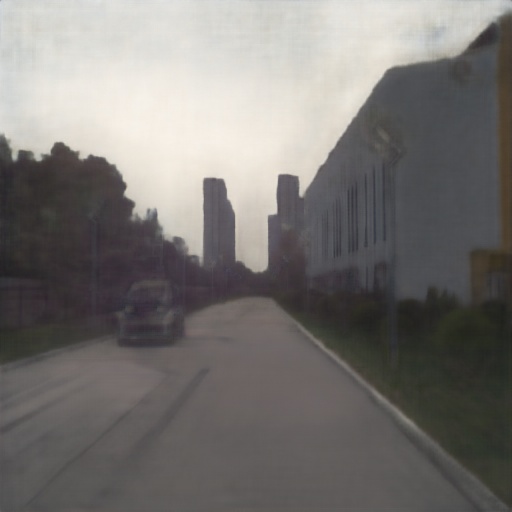} &
\includegraphics[width=\CellW,height=\CellH]{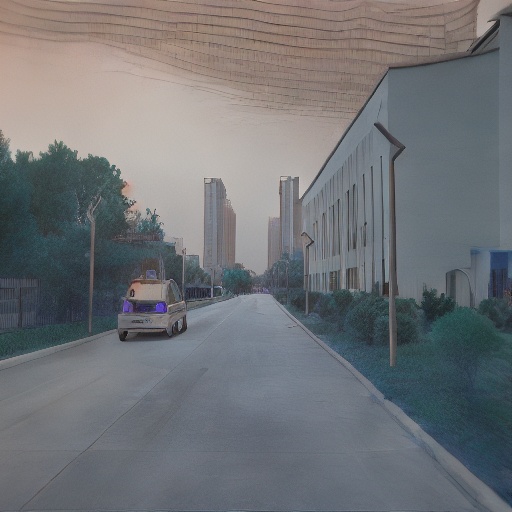} &
\includegraphics[width=\CellW,height=\CellH]{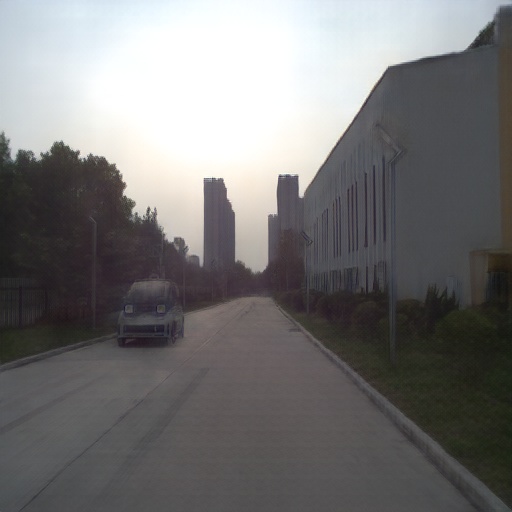} &
\includegraphics[width=\CellW,height=\CellH]{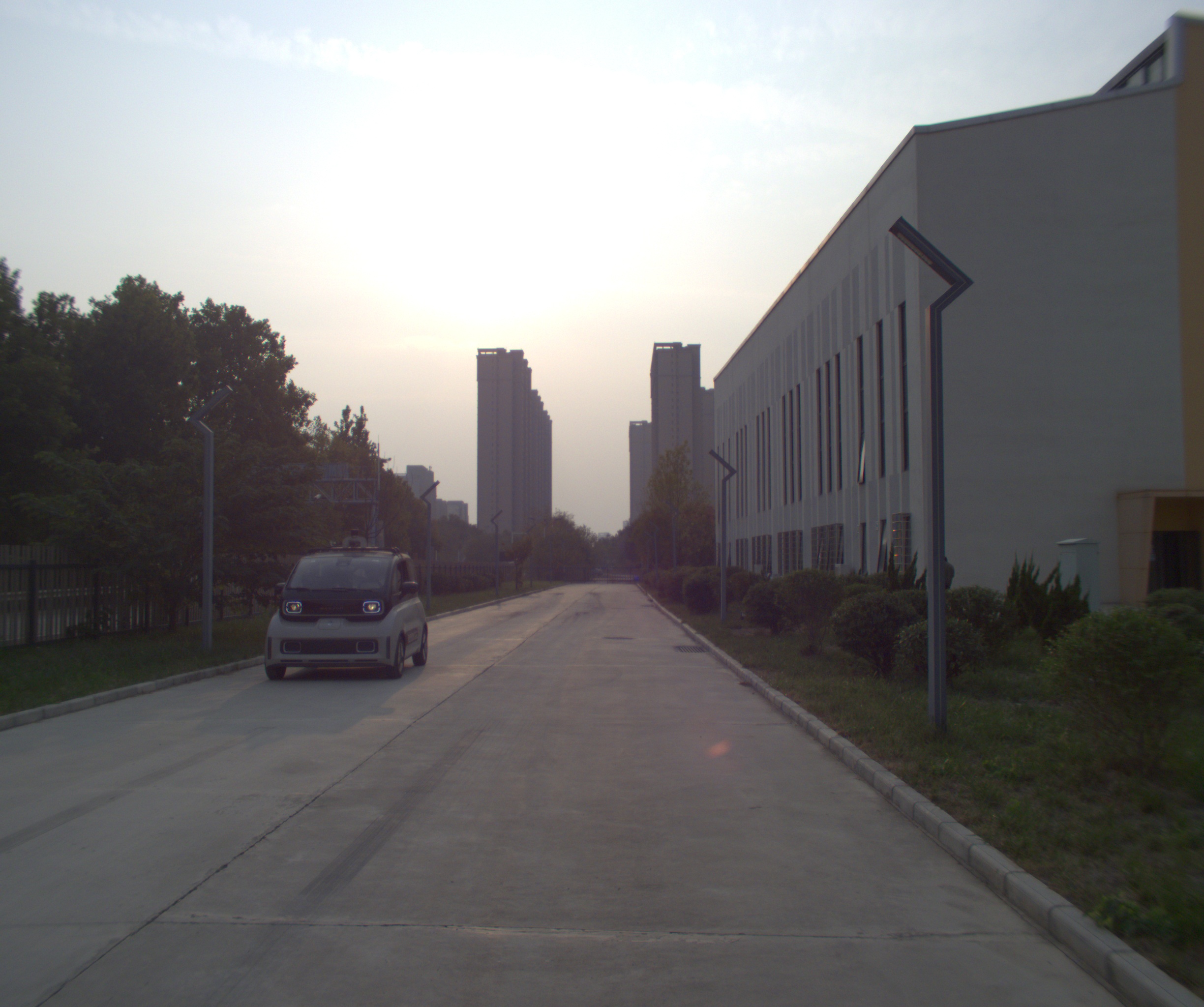} \\[-2pt]

\includegraphics[width=\CellW,height=\CellH]{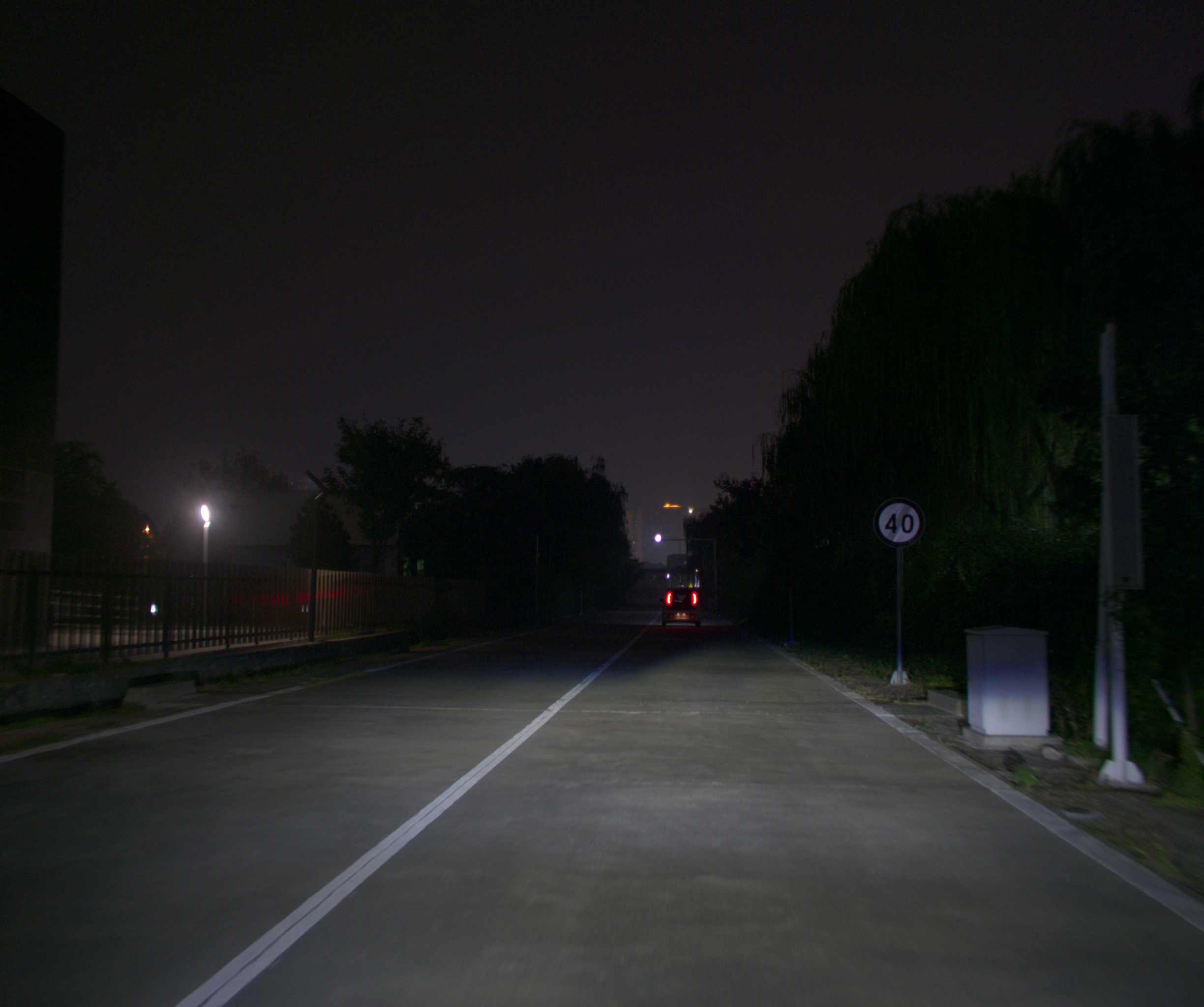} &
\includegraphics[width=\CellW,height=\CellH]{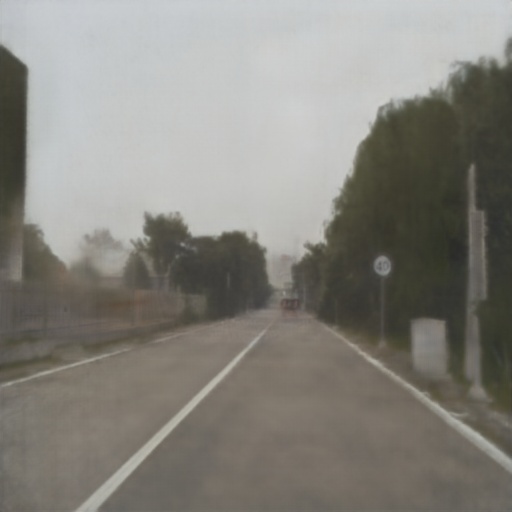} &
\includegraphics[width=\CellW,height=\CellH]{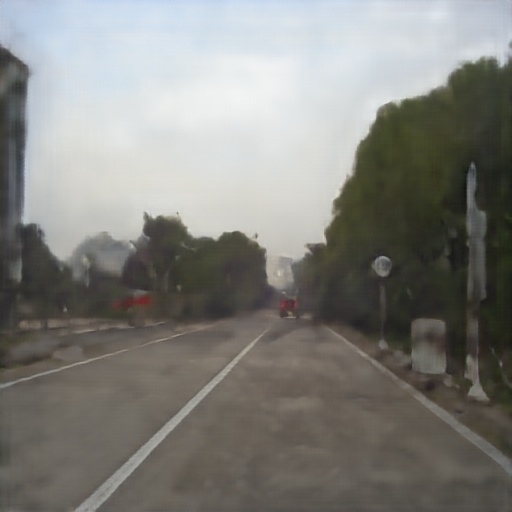} &
\includegraphics[width=\CellW,height=\CellH]{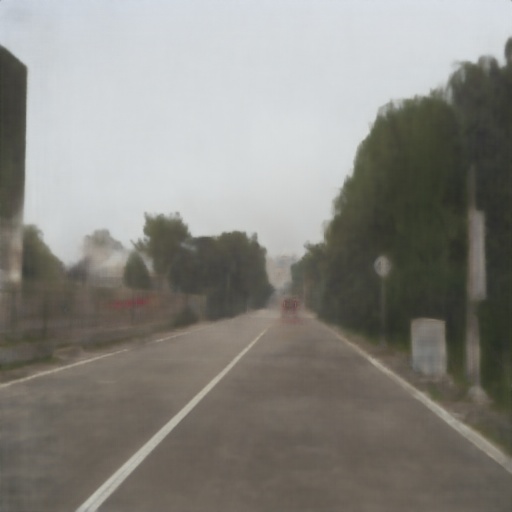} &
\includegraphics[width=\CellW,height=\CellH]{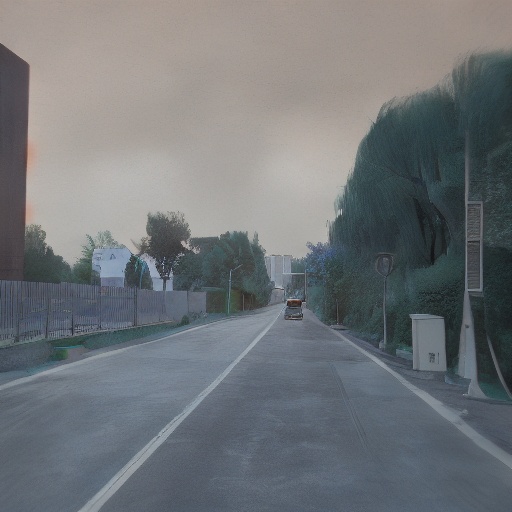} &
\includegraphics[width=\CellW,height=\CellH]{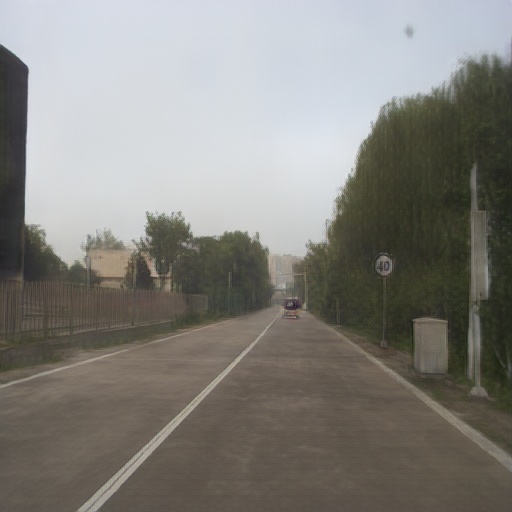} &
\includegraphics[width=\CellW,height=\CellH]{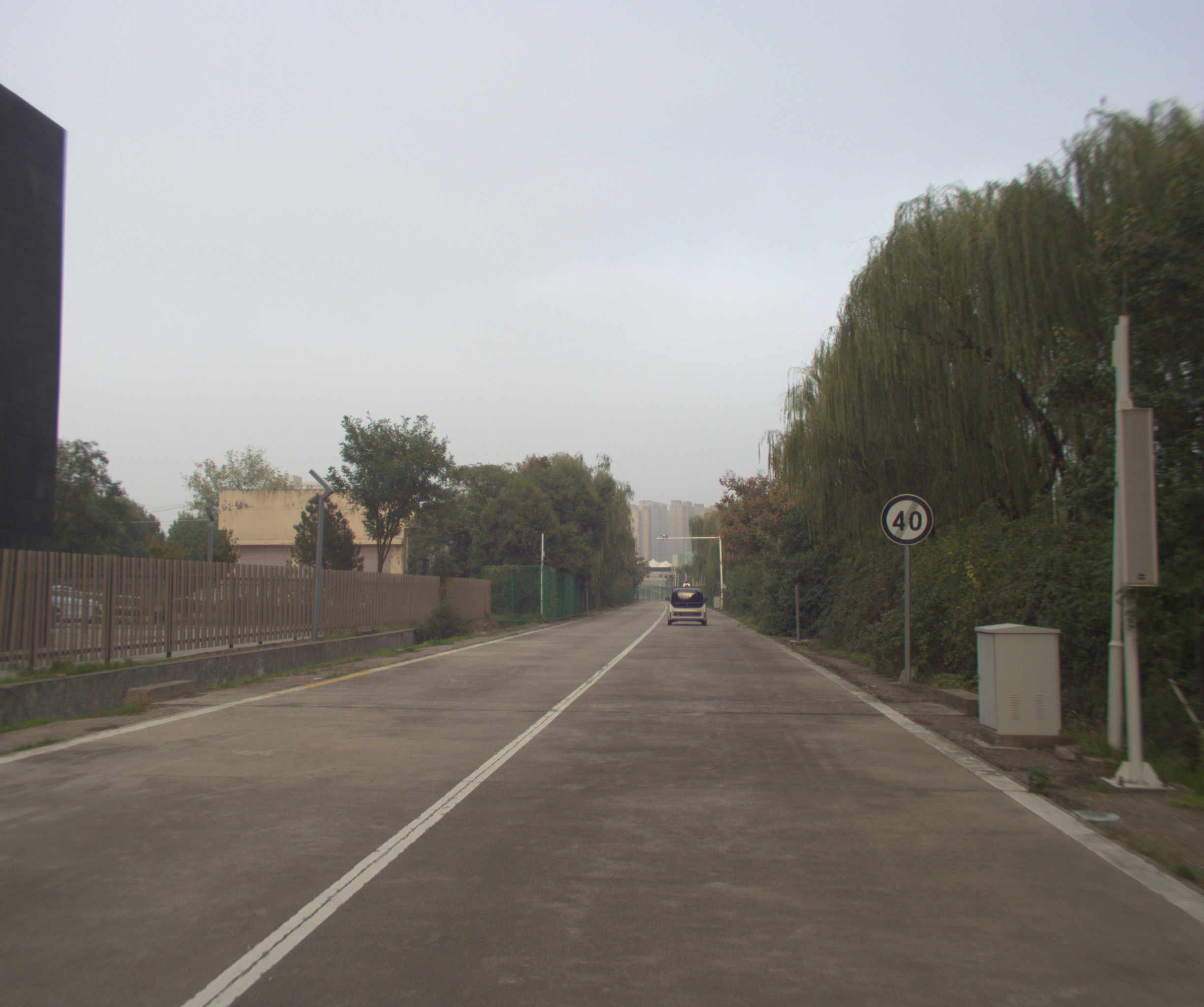}\\[-2pt]

\footnotesize (a) Night &
\footnotesize (b) LLFormer &
\footnotesize (c) SNR\mbox{-}SKF &
\footnotesize (d) Retinexformer &
\footnotesize (e) ControlNet &
\footnotesize (f) SNR\mbox{-}Aware &
\footnotesize (g) Day
\end{tabular}

\caption{\textbf{Visual comparison of nighttime image enhancement methods on the testing set of DarkDriving.}
The input is the nighttime image (a) and the outputs (b--f) are to approximate the corresponding daytime image (g).}
\vspace{-0.5em}
\label{fig:compare_low_light}
\end{figure*}

\subsection{Low-light Enhancement} 

\textbf{Experiment Setup.} For the low-light enhancement task, we evaluate five state-of-the-art methods on our DarkDriving  dataset, including SNR-Aware~\cite{xu2022snr}, SNR-SKF~\cite{wu2023learning}, LLFormer~\cite{wang2023ultra}, Retinexformer~\cite{cai2023retinexformer}, and ControlNet~\cite{zhang2023adding}. For each enhancement method, we use publicly available code and train each supervised learning-based method until convergence. Specifically, for ControlNet, we finetune the pre-trained Stable Diffusion v1-5 model~\cite{Rombach_2022_CVPR}. Before being fed into the network, all images are resized to 512×512 to maintain a consistent input format. We use three full-reference image quality metrics: PSNR, SSIM~\cite{wang2004image}, and LPIPS~\cite{zhang2018unreasonable} (Alex version), and four non-reference image quality assessment metrics: MUSIQ~\cite{ke2021musiq}, NIQE~\cite{mittal2012making}, HyperIQA~\cite{su2020blindly}, and CNNIQA~\cite{kang2014convolutional} to evaluate quantitative results.

\textbf{Experimental Results and Findings.} As shown in Table \ref{tab:low-light-enhancement} and Fig.~\ref{fig:compare_low_light}, SNR-Aware achieved high scores in both full-reference and no-reference metrics. ControlNet performs very well in non-reference metrics but sometimes shows some hallucination tradeoffs in full-reference metrics. Retinexformer also performs well. These results highlight two  key findings: 1) Low-light enhancement methods could \textit{significantly improve} the image quality of the original night image to approximate the corresponding daytime image (ground truth), leading to the promoted camera images for driving in the dark environment. 2) SNR-Aware shows the best balanced image quality after the low-light enhancement, whose strong SSIM and LPIPS results highlight its ability to enhance the dark images while maintaining structural integrity and perceptual quality.

\subsection{Generalized Low-light Enhancement} 
\textbf{Experimental Setup.} To verify the generalization capability  of our DarkDriving dataset, we test the trained  enhancement models on a different night driving dataset  nuScenes~\cite{caesar2020nuscenes}. First, we pre-train the SNR-Aware model (best performance in the previous enhancement experiment) with LOL-v2~\cite{wei2018deep} and DarkDriving datasets, respectively. To bridge the domain gap between datasets, following~\cite{li2024light}, we generate synthetic corresponding nighttime counterpart images of nuScenes daytime training set via digital image processing, then fine-tune the pre-trained SNR-Aware models with the synthetic night-day image pairs of nuScenes. Finally, we test the fine-tuned models pre-trained with these two datasets on the nighttime validation set of nuScenes.

\begin{figure}[htbp]
\centering
\setlength{\tabcolsep}{0.5pt}
\renewcommand{\arraystretch}{0.95}

\newcommand{\CellW}{.33\columnwidth}
\newcommand{\CellH}{.24\columnwidth}

\begin{tabular}{@{}ccc@{}}

\includegraphics[width=\CellW,height=\CellH]{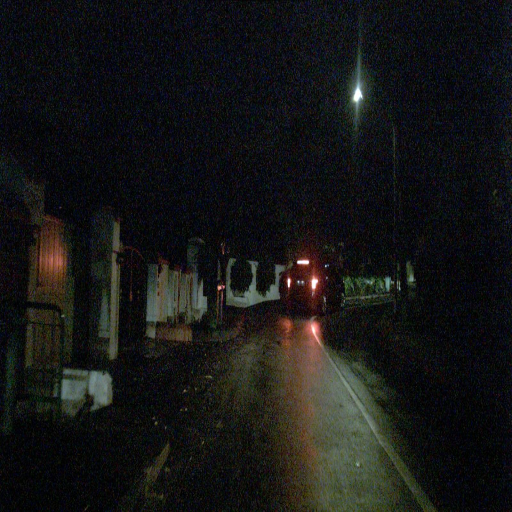} &
\includegraphics[width=\CellW,height=\CellH]{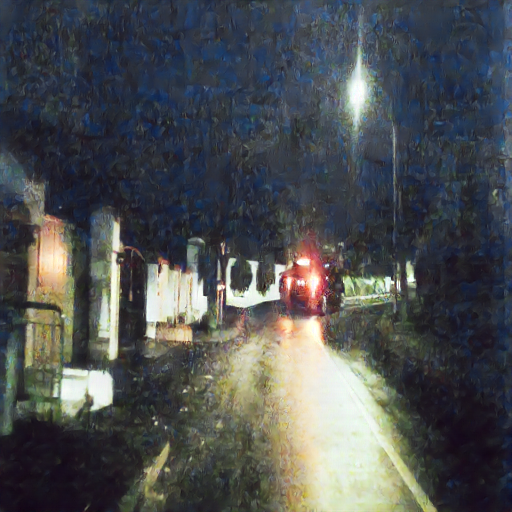} &
\includegraphics[width=\CellW,height=\CellH]{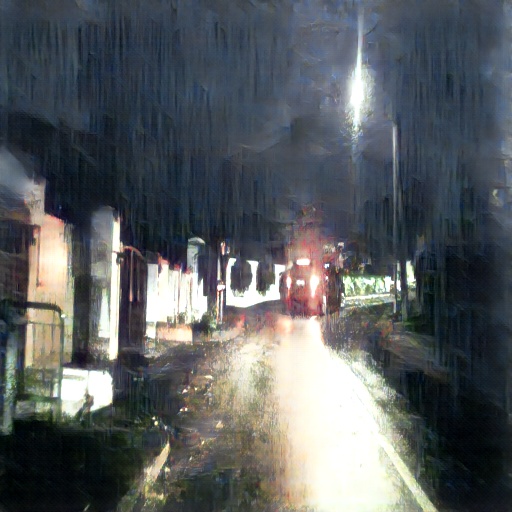} \\[-2pt]

\footnotesize Night &
\footnotesize LOL-v2 Pre-trained &
\footnotesize DarkDriving Pre-trained

\end{tabular}

\vspace{-0.5em}
\caption{\textbf{Visual Cross-dataset Enhancement on nuScenes}.}
\vspace{-1em}
\label{fig:nuscense_gen}
\end{figure}

\textbf{Experimental Results and Findings.} Table~\ref{tab:low-light-enhancement-nuscense} shows that the model pre-trained on our DarkDriving dataset produces better image quality compared to the model pre-trained on the LOL-v2 dataset, which is consistent in the visualized results of Fig.~\ref{fig:nuscense_gen}. These results confirm that our DarkDriving has better generalization than other datasets for low-light enhancement in the real-world dark driving environment. 





\begin{table}[htbp]
    \centering
    \caption{\textbf{Cross-dataset Enhancement.} Pre-trained on LOL-v2 or DarkDriving using SNR-Aware~\cite{xu2022snr}, Tested on the nighttime validation set of nuScenes.} 
    \label{tab:low-light-enhancement-nuscense}
    \resizebox{\columnwidth}{!}{
    \begin{tabular}{l|c|c|c|c}
        \toprule
        & \multicolumn{4}{c}{{Non-reference Metrics}} \\ 
        \cmidrule(lr){2-5} 
        {Pre-trained Dataset} & {MUSIQ↑} & {NIQE↓} & {HyperIQA↑} & {CNNIQA↑}\\ 
        \midrule
        LOL-v2-real   &  28.23 & 4.63 &  0.30 &  0.51 \\
        LOL-v2-synthetic   &  28.45 & 4.44 &  0.27 &  \textbf{0.54} \\
        DarkDriving   &  \textbf{35.83} &   \textbf{4.41} &  \textbf{0.31} &  0.53 \\
        \bottomrule
    \end{tabular}
    }
\end{table}


\subsection{Low-light Enhancement for 2D Detection}
\textbf{Experimental Setup.} 
We adopt YOLOv11~\cite{khanam2024yolov11} as the 2D detection backbone and consider two configurations: (1) a model pretrained on the COCO dataset~\cite{lin2014microsoft} without further adaptation, and (2) a variant finetuned on our DarkDriving daytime training set. Low-light enhancement methods are categorized into two groups: unsupervised and supervised. The raw nighttime image (“Night”) is used as the baseline. Evaluation is conducted on the dominant Car class following the YOLOv11 protocol, with AP$_{50}$ and AP$_{50-90}$ reported as the detection performance metrics.


\begin{table}[ht]
    \centering    
    \caption{\textbf{2D Detection on Darkdriving nighttime test set}. 
    $^*$: \emph{unsupervised} enhancement approaches. Other enhancement methods: \emph{supervised} and trained on the DarkDriving training set.}
    \label{tab:yolo11x-darkdriving}
    \resizebox{\columnwidth}{!}{
    \begin{tabular}{l|cc|cc}
        \toprule
        \multirow{2}{*}{Method} & \multicolumn{2}{c|}{YOLOv11-pretrained} & \multicolumn{2}{c}{YOLOv11-finetune} \\
        \cmidrule(lr){2-3} \cmidrule(lr){4-5}
         & {AP$_{50}$↑} & {AP$_{50-90}$↑} & {AP$_{50}$↑} & {AP$_{50-90}$↑} \\
        \midrule
        Night           & \underline{0.316} & \underline{0.178} & 0.166 & 0.083 \\ 
        \midrule
        CLIP-LIT$^*$~\cite{liang2023iterative}     & 0.309 & 0.174 & 0.183 & 0.093\\
        EnlightenGAN$^*$~\cite{jiang2021enlightengan} & 0.291 & 0.165 & 0.093 & 0.048 \\
        LLFormer~\cite{wang2023ultra} & 0.161 & 0.090 & 0.167 & 0.071 \\
        Retinexformer~\cite{cai2023retinexformer} & 0.173 & 0.090 & \underline{0.215} & \underline{0.105} \\
        LightTheNight~\cite{li2024light} & 0.227 & 0.093 & 0.141 & 0.053 \\
        SNR-Aware~\cite{xu2022snr} & \textbf{0.363} & \textbf{0.187} & \textbf{0.550} & \textbf{0.265} \\
        \bottomrule
    \end{tabular}
    }
\end{table}

\begin{figure}[t]
\centering
\setlength{\tabcolsep}{0.5pt}
\renewcommand{\arraystretch}{0.95}

\newcommand{\CellW}{.245\columnwidth}
\newcommand{\CellH}{.245\columnwidth}

\begin{tabular}{@{}cccc@{}}

\includegraphics[width=\CellW,height=\CellH]{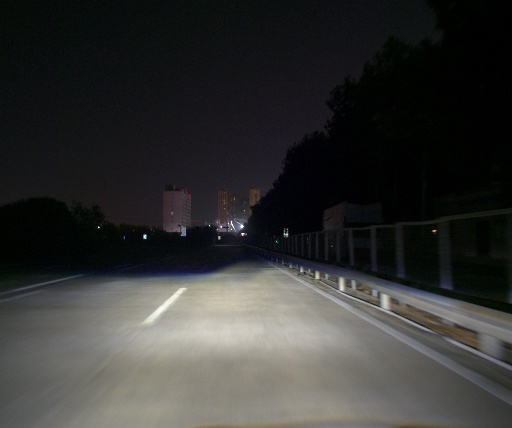} &
\includegraphics[width=\CellW,height=\CellH]{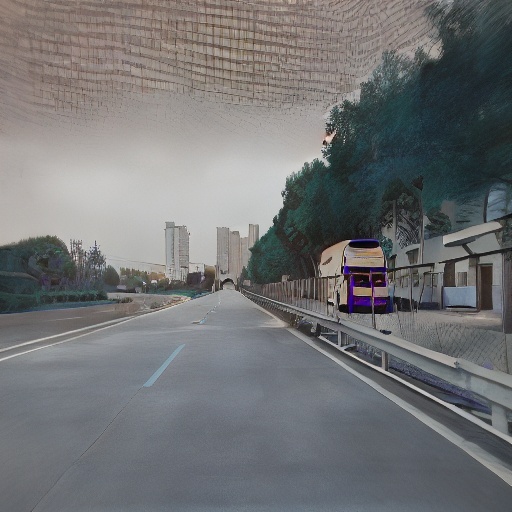} &
\includegraphics[width=\CellW,height=\CellH]{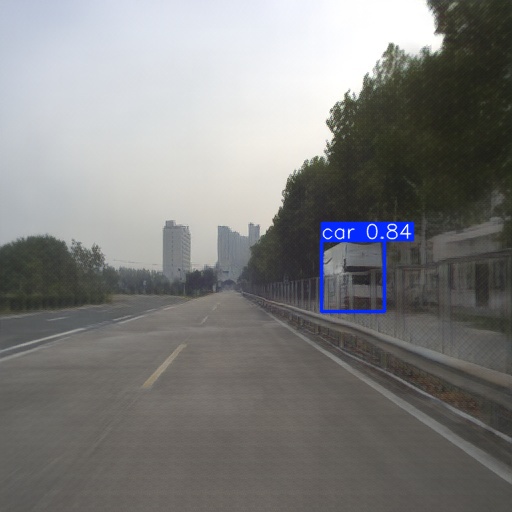} &
\includegraphics[width=\CellW,height=\CellH]{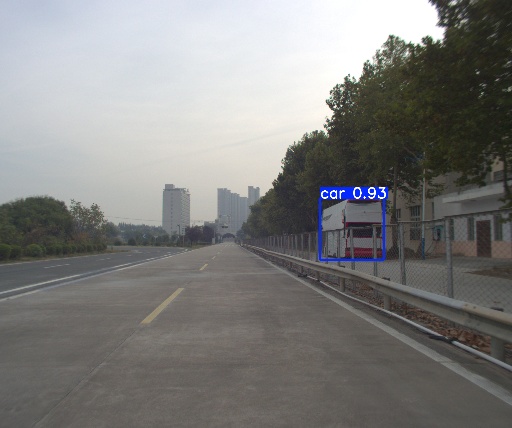} \\[-2pt]

\includegraphics[width=\CellW,height=\CellH]{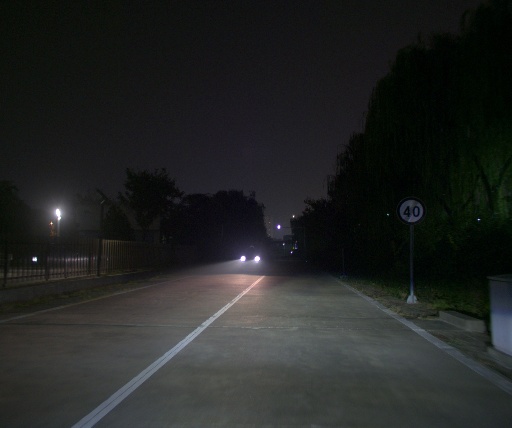} &
\includegraphics[width=\CellW,height=\CellH]{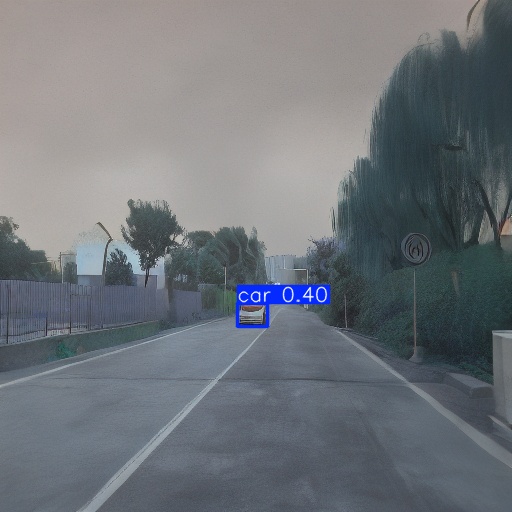} &
\includegraphics[width=\CellW,height=\CellH]{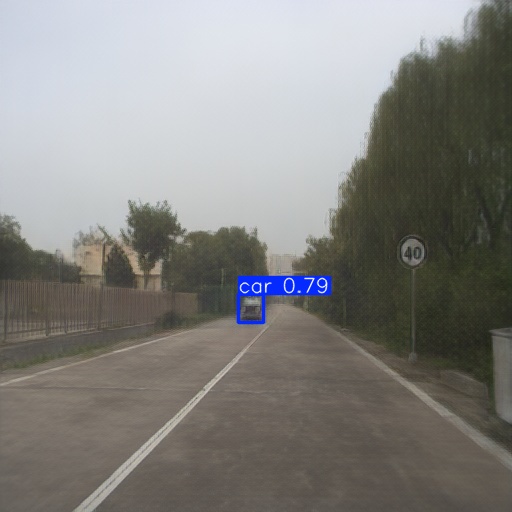} &
\includegraphics[width=\CellW,height=\CellH]{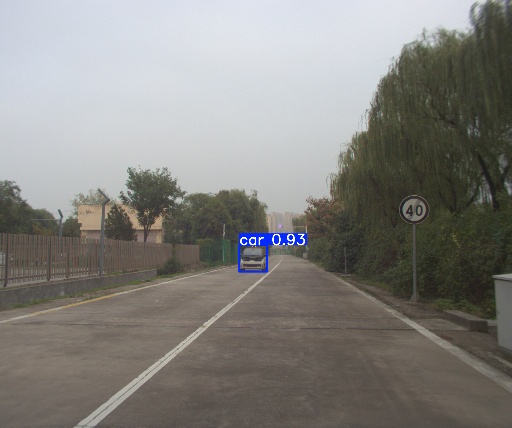} \\[-2pt]

\footnotesize (a) Night &
\footnotesize (b) LightTheNight &
\footnotesize (c) SNR-Aware &
\footnotesize (d) Day

\end{tabular}

\vspace{-0.5em}
\caption{\textbf{2D detection errors on nighttime images can be reduced after low-light enhancement. (a) and (d) are night-day paired data of DarkDriving dataset.}}
\vspace{-1em}
\label{fig:compare_low_light_2d}
\end{figure}

\textbf{Experimental Results and Findings.} 
Table~\ref{tab:yolo11x-darkdriving} shows that with the COCO-pretrained detector, only SNR-Aware improves nighttime detection, consistent with Table~\ref{tab:low-light-enhancement} where it achieves the best image quality. DarkDriving daytime data-finetuned detector does not perform well on the DarkDriving nighttime data (0.166 AP$_{50}$), while  DarkDriving-trained SNR-Aware enhancement restores performance to 0.550 AP$_{50}$. Figure~\ref{fig:compare_low_light_2d} further confirms that missing detection in dark can be found after enhancement. These results demonstrate that paired enhancement on DarkDriving is effective in bridging the day–night gap for robust 2D detection.


\subsection{Low-light Enhancement for 3D Detection}

\textbf{Experimental Setup.} We train the supervised SNR-Aware model on the paired DarkDriving dataset and then evaluate it on the nuScenes 3D detection benchmark. Our experimental setup and comparison methods follow the protocol in LightTheNight~\cite{li2024light}. During testing, we applied low-light enhancement exclusively to the front camera images of the nuScenes nighttime validation set, while keeping the other five camera views unchanged. The evaluation metrics are Average Precision (AP), Average Translation Error (ATE), and Average Orientation Error (AOE).


\begin{table}[ht]
    \centering
    \caption{\textbf{3D Detection on nuScenes nighttime validation set}. 
    BEVDepth (RGB-Only)~\cite{li2023bevdepth} and CRN (RGB+Radar)~\cite{kim2023crn} are trained on nuScenes daytime training set. 
    $^*$: \emph{unsupervised} enhancement approaches. LightTheNight: trained on nuScenes. SNR-Aware: pre-trained on DarkDriving training set (same as that in Table~\ref{tab:low-light-enhancement-nuscense}). }
    \label{tab:low-light-enhancement-downstream}
    \resizebox{\columnwidth}{!}{
    \begin{tabular}{l|c|c|c|c|c|c}
        \toprule
        & \multicolumn{3}{c|}{{BEVDepth (RGB)}} & \multicolumn{3}{c}{{CRN (RGB+Radar)}} \\ 
        \cmidrule(lr){2-4} \cmidrule(lr){5-7}
        {Method} & {AP↑} & {ATE↓} & {AOE↓} & {AP↑} & {ATE↓}  & {AOE↓} \\ 
        \midrule
        Night           & 0.134 & 0.787  & 0.957 &  0.435 & 0.474 &  0.597 \\ 
        \midrule
        CLIP-LIT$^*$~\cite{liang2023iterative} & 0.131 & 0.791 & 0.972  & 0.430 & 0.475 & 0.607 \\
        EnlightenGAN$^*$~\cite{jiang2021enlightengan} & 0.138 & \underline{0.786} & \underline{0.948} & 0.436 & \underline{0.473} & 0.585 \\
        LightTheNight~\cite{li2024light} & \textbf{0.176} & \textbf{0.774} & 1.108 & \textbf{0.485} & 0.478 & \textbf{0.464} \\
        SNR-Aware~\cite{xu2022snr} & \underline{0.148} & \underline{0.786} & \textbf{0.911} & \underline{0.481} & \textbf{0.468} & \underline{0.575} \\
        \bottomrule
    \end{tabular}
    }
\end{table}

\textbf{Experimental Results and Findings.} As shown in Table~\ref{tab:low-light-enhancement-downstream}, compared to directly using nighttime images, the low-light enhancement model SNR-Aware simply pre-trained with our DarkDriving dataset could improve 3D detection on nuScenes and get comparable performance with the state-of-the-art method LightTheNight~\cite{li2024light}  intensively trained on nuScenes, which indicates that our DarkDriving can be generalized to promote detection in other low-light driving environment, such as nuScenes.

%% file: sec_camera/5_conclusion.tex
\section{Conclusion}\label{sec:con}
This paper introduces the first real-world benchmark dataset, DarkDriving, to advance research on low-light enhancement for autonomous driving in nighttime. By designing a new Trajectory Tracking based Pose Matching (TTPM) method for the large closed driving test field, the day-night image pairs with precise location and spatial-content alignments (error: centimeters) are  collected automatically. Using DarkDriving, we study four perception tasks: low-light enhancement, generalized low-light enhancement, and low-light enhancement for 2D and 3D detection. Our DarkDriving work established a comprehensive benchmark of evaluating low-light enhancement for autonomous driving, and it can be generalized to benefit low-light enhancement and enhance detection on other datasets, such as nuScenes.